\documentclass[final,12pt]{clear2026} 

\title[Causal Process Models]{Causal Process Models:
Reframing Dynamic Causal Graph Discovery as a Reinforcement Learning Problem}
\usepackage{times}
\usepackage{booktabs}
\usepackage{hyperref}
\usepackage{url}

\usepackage{amssymb}            
\usepackage{mathtools}          
\usepackage{mathrsfs}           
\usepackage{dsfont}
\usepackage{float}
\usepackage{graphicx}           
\usepackage{subcaption}         
\usepackage[space]{grffile}     
\usepackage{url}                
\usepackage{lipsum}             
\usepackage{fontawesome5}

\usepackage{microtype}      
\usepackage{xcolor}         
\usepackage{amsmath}
\usepackage{mathtools}
\usepackage{tikz-cd}
\usepackage{tikz}
\usetikzlibrary{cd, calc} 

\definecolor{magentaObj}{HTML}{D00073}
\definecolor{violetFor}{HTML}{4000FF}
\definecolor{objBlue}{HTML}{0052f8}
\definecolor{outlineOrange}{HTML}{e29800}

\hypersetup{
    colorlinks,
    linkcolor=magentaObj,
    citecolor=violetFor,
    urlcolor=objBlue
}

\clearauthor{%
 \Name{Turan Orujlu} \Email{turan.orujlu@tuebingen.mpg.de}\\
 \addr 
University of T{\"u}bingen \& MPI for Biological Cybernetics
 \AND
 \Name{Christian Gumbsch} \\
 \addr 
University of Amsterdam 
 \AND
 \Name{Martin V. Butz} \\
 \addr 
University of T{\"u}bingen 
 \AND
 \Name{Charley M. Wu} \\
 \addr 
TU Darmstadt \& Hessian.AI 
}

\begin{document}

\maketitle

\begin{abstract}%
  Most neural models of causality assume static causal graphs, failing to capture the dynamic and sparse nature of physical interactions where causal relationships emerge and dissolve over time.
 We introduce the Causal Process Framework and its neural implementation, Causal Process Models (CPMs), for learning sparse, time-varying causal graphs from visual observations. Unlike traditional approaches that maintain dense connectivity, our model explicitly constructs causal edges only when objects actively interact, dramatically improving both interpretability and computational efficiency.
 We achieve this by casting dynamic interaction-graph construction for world modeling as a multi-agent reinforcement learning problem, where specialized agents sequentially decide which objects are causally connected at each timestep.
 Our key innovation is a structured representation that factorizes object and force vectors along three learned dimensions (mutability, causal relevance, and control relevance), enabling the automatic discovery of semantically meaningful encodings.
 We demonstrate that a CPM significantly outperforms dense graph baselines on physical prediction tasks, particularly for longer horizons and varying object counts.%
\end{abstract}


\section{Introduction}
\label{sec:introduction}

Causality plays a fundamental role in building intelligent systems capable of physical reasoning \citep{Gerstenberg2020ACS}. Explicitly modeling causal relationships is increasingly recognized to be crucial for developing robust, generalizable, and interpretable neural network models capable of accurate prediction and effective intervention \citep{DBLP:conf/nips/XiaLBB21}. 
Despite their black-box nature, models such as transformers have demonstrated surprising capacity for causal reasoning \citep{DBLP:conf/icml/NichaniDL24, DBLP:conf/nips/ShouBGSHB23, DBLP:conf/icml/MelnychukFF22, dettki2025large}. One explanation posits that this is possible due to the attention mechanism forming implicit causal edges between tokens \citep{DBLP:conf/nips/VaswaniSPUJGKP17, DBLP:conf/nips/RohekarGN23}. 
However, recent work has highlighted a phenomenon known as \emph{over-squashing}, in which the attention mechanism (and related message-passing mechanisms in Graph Neural Networks) loses sensitivity to individual tokens or nodes \citep{DBLP:journals/corr/abs-2406-04267, DBLP:conf/iclr/0002Y21, DBLP:conf/iclr/BarberoVSBG24, DBLP:conf/icml/GiovanniGBLLB23, DBLP:journals/tmlr/GiovanniRBDLMV24, DBLP:conf/iclr/ToppingGC0B22, DBLP:journals/tnn/ScarselliGTHM09, DBLP:journals/corr/battaglia_gnn}. This compression of information in transformer models can sever causal chains, thus limiting the effectiveness of causal inference.

\begin{figure*}[t]
    \centering
    \begin{minipage}[b]{0.32\linewidth}
        \centering
        \includegraphics[width=\linewidth]{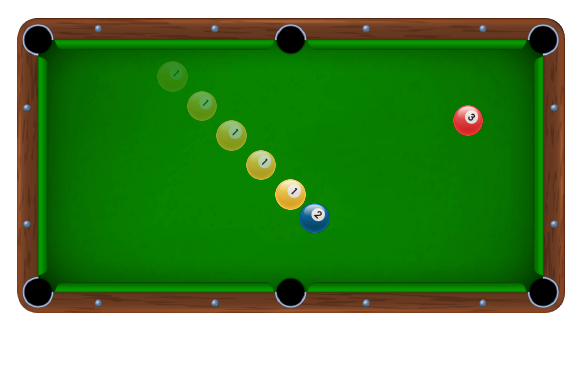}
        \\ (a) example trajectory
    \end{minipage}
    \hfill
    \begin{minipage}[b]{0.28\linewidth}
        \centering
        \includegraphics[width=\linewidth]{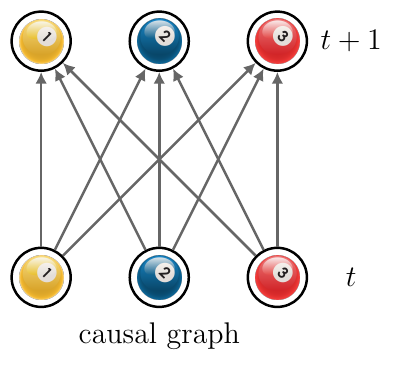}
        \\ (b) static causal graph
    \end{minipage}
    \hfill
    \begin{minipage}[b]{0.35\linewidth}
        \centering
        \includegraphics[width=\linewidth]{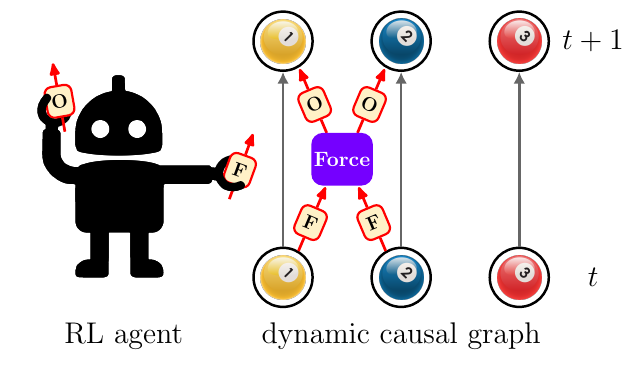}
        \\ (c) CPM's dynamic causal graph
    \end{minipage}

    \vspace{0.2cm} 
    \caption{\textbf{Dynamic Causality}: (a) In many physical domains, such as a game of billiards, objects interact only sparsely. (b) Static causal graphs must encode \textit{all possible} interactions, resulting in dense connectivity that fails to capture this local sparsity. (c) In a Causal Process Model (CPM), an RL agent dynamically constructs a causal graph by connecting forces and objects through process blocks, yielding a sparse, dynamic causal graph that reflects the actual interactions at each timestep.}
    \label{fig:example}
\end{figure*}

In contrast, graphical causal models, such as Pearl's Structural Causal Models \citep[SCMs;][]{Pearl2009CausalII}, explicitly encode causal relationships and thus preserve perfect causal connectivity by design. 
Yet, a key challenge for SCMs is \emph{causal discovery} \citep{scholkopf2021toward}: inferring the causal graph from data. 
Most existing approaches assume access to a complete dataset and construct a static causal graph, e.g., for all possible interactions of three billiard balls a dense graph is necessary (see Fig.~\ref{fig:example}a).
This assumption is misaligned with the nature of physical environments, where causal influence is typically local in space and sparse in time \citep{Butz:2017b, pitis2020counterfactual, seitzer2021causal, gatel0rd, DBLP:journals/corr/lange_kording}.
For instance, objects may only interact upon contact.
Recent work has therefore emphasized the importance of \emph{local causal models} \citep{pitis2020counterfactual, seitzer2021causal, urpi2024causal, spartan, DBLP:conf/iclr/WilligTBSDK25} that explain causal connections through the sparsest possible graph, which changes dynamically over time. 

Our work aims to bridge these areas by proposing a novel causal framework tailored to capture the dynamics of physical object interactions. We propose \textbf{Causal Process Models} (CPMs), as a neural implementation of this framework \textbf{casting the construction of sparse dynamic causal graphs as a sequential reinforcement learning (RL) problem}. Instead of relying on dense message passing (e.g., soft attention or standard GNNs, Fig.~\ref{fig:example}b), CPMs use RL agents to dynamically determine all-or-nothing connections between entities (Fig.~\ref{fig:example}c).
This allows the model to adaptively control connectivity based on the input, avoiding the over-squashing problem and enabling more efficient and interpretable causal reasoning. 

Our novel causal framework is designed specifically for modeling the dynamics of physical object interactions, aiming to synthesize the formal rigor of \textit{static dependency} theories, e.g. Pearl's do-calculus \citep{Pearl2009CausalII}, with the intuitive strengths of \textit{process-based} accounts \citep[][see Section~\ref{sec:related_work} below]{Russell1948-RUSHKI-5, Salmon1984-SALSEA, Skyrms1981-SKYCN, Dowe2000-DOWPC-2}. Our approach explicitly addresses the limitations of Pearlian SCMs by enabling the construction of sparse, time-varying causal graphs that reflect only the active interactions between objects. When modeling two colliding balls for instance, our framework only instantiates a direct causal link between the balls upon contact, for the transfer of momentum, while leaving them causally disconnected otherwise. 
This yields a computationally efficient model, only scaling with actual rather than all potential interactions, and one that is highly interpretable since the causal graph mirrors intuitive physical processes. 

Our main contributions are: 1) We formalize a \textit{Causal Process Framework} (CPF) for local causal modeling in physical environments. 2) We implement this in a neural architecture as a \textit{Causal Process Model} (CPM) to dynamically infer sparse, time-varying causal graphs by framing edge selection as an RL problem. 3) We apply our CPM to physical interaction scenarios, demonstrating superior performance, interpretability, and scalability compared to densely connected models.

\begin{figure}[t]
    \centering
    \includegraphics[width=0.95\textwidth]{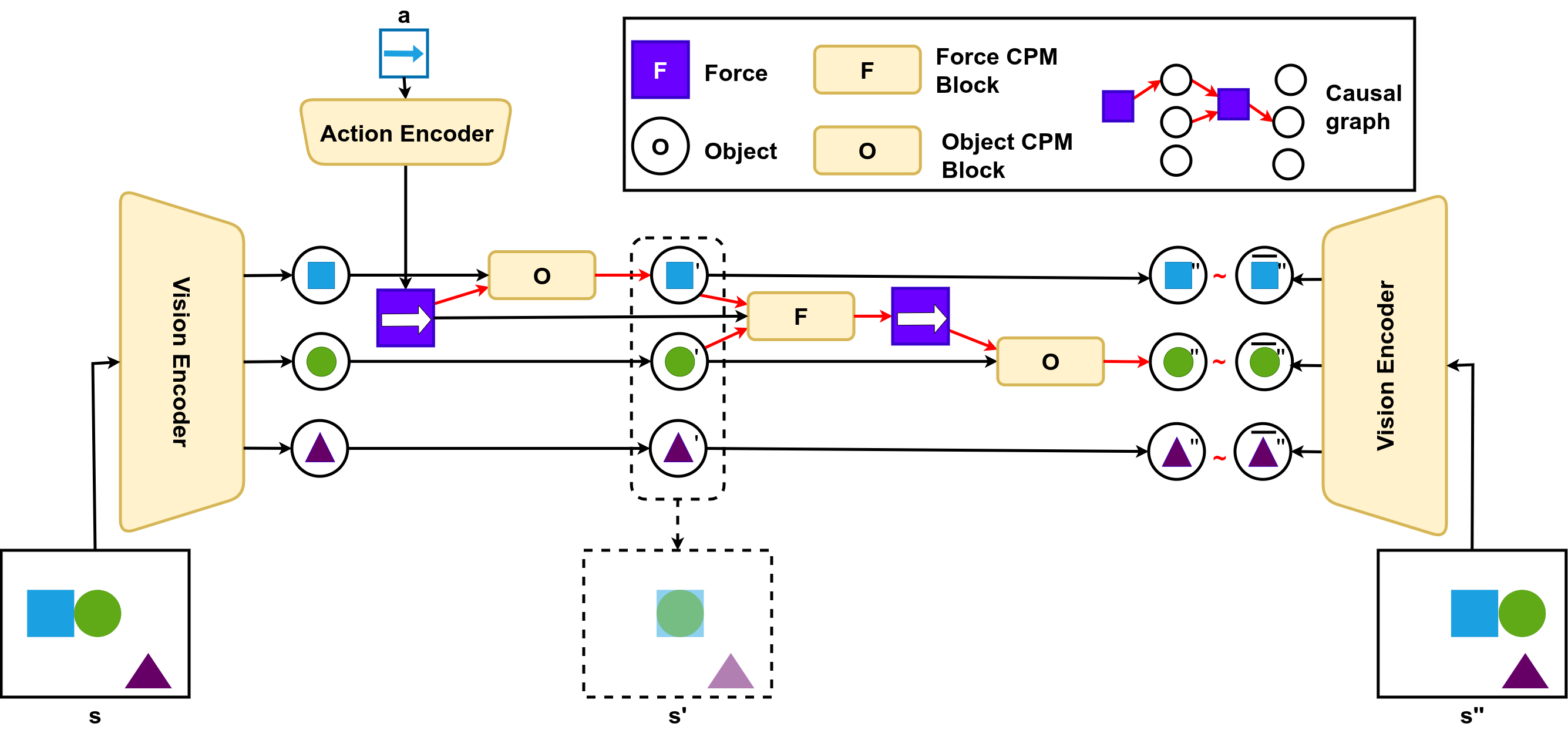}
    \vspace{-1.0em}
    \caption{\textbf{Model Overview}: Our model has three components: a vision encoder, an action encoder, and a transition function. The transition function is an implementation of a \emph{Causal Process Model}. The state is factorized into distinct object representations, actions are mapped to force representations that act as causal interventions, and the directed edges are causal. \url{https://github.com/torujlu/CausalProcessModels}}
    \label{fig:edges-intervention-model}
    \vspace{-1.5em}
\end{figure}

\section{Related Work}
\label{sec:related_work}

\subsection{Causal frameworks}
\label{sec:causal frameworks}

Pearl's (\citeyear{Pearl2009CausalII}) framework of Structural Causal Models (SCMs) is a dominant approach to causal modeling, by representing causal relationships using directed acyclic graphs (DAGs). An SCM can be described as a tuple $\mathfrak{C}:=\left(\mathbf{S}, \mathbb{P}(\mathbf{U})\right)$ where $\mathbb{P}$ is a distribution over the exogenous variables $\mathbf{U}$ (i.e., variables external to the system and not caused by any variable within it) and $\mathbf{S}$ is a collection of structural equations of the form:

\begin{align*}
    V_i &= f_{V_i}(\mathbf{Pa}_{V_i}, \mathbf{U}_{V_i}).
\end{align*}

Each endogenous variable $V_i$ is determined by a function of its parent variables $\mathbf{Pa}_{V_i}$ (i.e., other variables in the system that directly influence $V_i$) and its associated exogenous noise term $\mathbf{U}_{V_i}$.

While successful in many domains, standard SCMs require extensions to handle systems characterized by dynamic object interactions; without such extensions, they fail to adequately capture the temporal and structural intricacies of such systems \citep{Rubenstein2016FromDO, Weber2016, pmlr-v115-blom20a, Boeken2024DynamicSC}.
Consider the simple scenario of two colliding balls shown in Fig.~\ref{fig:example}a.
Representing this within a traditional SCM framework often requires specifying potential causal links between all properties of all objects at all relevant timescales. This leads to densely connected causal graphs (Fig.~\ref{fig:example}b), with the number of causal edges scaling quadratically with time, even when interactions are sparse in reality. Such dense representations suffer from high computational costs for inference and learning, and crucially, obscure the underlying causal structure, hindering interpretability. Thus, a core challenge is to adapt standard SCMs to dynamically represent only the relevant interactions as they occur, rather than needing to specify all potential dependencies. 

Recognizing these limitations, other lines of research offer valuable perspectives, often aligning closely with \textit{causal process theories} \citep{Russell1948-RUSHKI-5, Salmon1984-SALSEA, Skyrms1981-SKYCN, Dowe2000-DOWPC-2}.
Research in cognitive science, such as \citet{Gerstenberg2020ACS}'s counterfactual simulation models, leverage simulation to assess causality and responsibility in physical events, capturing process-like intuitions. Furthermore, philosophical inquiries into causal processes provide rich conceptual foundations, distinguishing \textit{causal} processes from \textit{pseudo}-processes by focusing on mechanisms like causal lines \citep{Russell1948-RUSHKI-5}, defining causality in ontological terms \citep{Salmon1984-SALSEA},  or using conserved quantities \citep{Skyrms1981-SKYCN, Dowe2000-DOWPC-2}. However, this philosophical tradition lacks the computational formalism required for direct implementation in ML systems.
Our Causal Process Framework bridges this gap by providing a computationally tractable formalism that integrates process-based intuitions with graphical causal models, enabling dynamic and sparse representations suitable for learning from visual data in physical environments.

\subsection{Neural Causal Models}
\label{sec:causal_models}

While philosophical causal process theories offer intuitive insights into dynamic physical interactions, their abstract nature limits direct application in scalable machine learning systems. To operationalize these ideas computationally, researchers have sought to integrate causal process intuitions with neural architectures, particularly by embedding SCMs into deep learning frameworks.
Previous attempts to reconcile deep learning with SCMs have resulted in Neural Causal Models (NCMs), which model $f_{V_i}$ as feedforward neural nets parametrized by $\theta_{V_i}$ \citep{DBLP:conf/nips/XiaLBB21}. Yet this solution still suffers from the disadvantage of needing to train arbitrarily many feedforward neural networks for each node across time. To address this parameter explosion, \citet{DBLP:journals/corr/ncm-gnn} have tried to theoretically quantify the capacity for GNNs to implement SCMs, but are restricted to the assumption of static causal graph. In contrast, \citet{DBLP:conf/icml/MelnychukFF22} designed a Causal Transformer that incorporates temporal dynamics to infer causality over time, yet is still unable to yield interpretable graph representations.
This limitation arises from its reliance on the potential outcomes framework \citep{10.1214/aos/1176344064, inbook}, which focuses on estimating counterfactual outcomes without explicitly representing causal relationships as graphs, thus making it less suitable for discovering and utilizing sparse, time-varying structures.

\subsection{Causal Reinforcement Learning}
\label{sec:causal_rl}

\citet{DBLP:conf/iclr/BuesingWZHRGL19} have tried to take advantage of the Pearlian causality framework by reformulating the MDP graph as an SCM using which they designed a counterfactually-guided policy search. A similar approach has been pursued by \citet{DBLP:journals/tmlr/GasseGGO23} in which they draw parallels between confounding variables and offline RL. Neither of these approaches factors the MDP state space into distinct object-centric nodes and their causal relations, instead focusing on the aforementioned inherent causality of the MDP structure as suggested by \citet{bareinboim2021introduction}.

\subsection{RL for causal discovery}

Several works formulate causal structure learning as a reinforcement-learning problem, where the policy searches over graph structures to optimize a dataset-level score for a \emph{single} global causal graph (e.g., score-based objectives over tabular variables). \citet{DBLP:conf/iclr/ZhuNC20} and \citet{DBLP:journals/tmlr/Duong0HN25} use RL to explore the combinatorial space of adjacency structures, while CORE considers an active discovery setting with interventions \citep{DBLP:conf/atal/SauterBAP24}. In contrast, our CPM uses RL \emph{within} a predictive world model to construct sparse, time-varying interaction graphs conditioned on the current visual state. Our output is not a single global DAG but a per-timestep interaction structure used for forward prediction and downstream model-based control.

\section{Causal Process Framework}

Pearl's structural causal models (SCMs) and do-calculus \citep{Pearl2009CausalII} provide a powerful foundation for causal reasoning.
However, without extensions, it is not straightforward to apply SCM to dynamic physical systems requiring object-centric representations and real-time causal interactions (see Appendix \ref{sec:object-centric_causal_dynamics} for a demonstration).
Prior approaches \citep{DBLP:conf/iclr/BuesingWZHRGL19, DBLP:journals/tmlr/GasseGGO23} have attempted to bridge model-based RL and causality by representing the full Markov Decision Process (MDP) state $s^t$ using a single node and modeling actions as direct interventions in a static causal graph. However, this approach is limited because it circumvents the problem of inferring the causal structure that generates the underlying environment dynamics \citep[i.e., the causal context;][]{BUTZ2025105948}, and focuses only on the causal implications of action sequences.

\subsection{Causal Process Models (CPMs)}

To address the inability of SCMs to capture sparse, time-varying interactions, the computational burden of dense connectivity, and the loss of causal information in over-squashed message-passing, we introduce Causal Process Models (CPMs). CPMs dynamically construct sparse causal graphs that represent only active interactions, enabling both computational efficiency and interpretable causal structure in physical environments.

We adopt an \textit{object-centric factorization} of states, in which physical objects are represented separately as object nodes $\mathcal{O} = \{O_1, O_2, ..., O_N\}$ (e.g., balls) and interactions between objects are represented as force nodes $\mathcal{F} = \{F_1, F_2, ..., F_M\}$ (e.g., collisions). At each timestep $t$, we have object states $\mathcal{O}^t = \{O^t_1, ..., O^t_N\}$ and force states $\mathcal{F}^t = \{F^t_1, ..., F^t_M\}$.  The key insight is that not all objects interact at all times, hence we need to dynamically determine which causal edges are active.

\subsubsection{Dynamic Causal Graph Construction}
To dynamically determine active causal edges in the graph, CPMs employ two types of specialized controller functions: \textit{interaction scope controllers} $\rho^t_{\mathcal{O}}$ determine which objects interact, that is, exchange forces (e.g., based on spatial proximity), while \textit{effect attribution controllers} $\rho^t_{\mathcal{O}\leftrightarrow\mathcal{F}}$ determine how objects are affected by these interacting forces. 

Formally, each controller outputs probabilistic distributions over possible edge subsets at each timestep. The interaction scope controllers $\rho^t_{\mathcal{O}}$ define a distribution over edge sets $J^t \subseteq \mathcal{O}^t \times \mathcal{F}^t$ conditioned on current object states $\mathcal{O}^t$ , yielding $J^t \sim \rho^t_{\mathcal{O}}(\cdot \mid \mathcal{O}^t)$.
Similarly, the effect attribution controllers $\rho^t_{\mathcal{O}\leftrightarrow\mathcal{F}}$ define a distribution over edge sets $I^t \subseteq \mathcal{F}^t \times \mathcal{O}^{t+1}$ conditioned on current object and force states $(\mathcal{O}^t, \mathcal{F}^t)$, yielding $I^t \sim \rho^t_{\mathcal{O}\leftrightarrow\mathcal{F}}(\cdot \mid \mathcal{O}^t, \mathcal{F}^t)$.

Within this framework, state evolutions are governed by object and force update functions $f_O$ and $f_F$, which propagate information along the dynamically selected causal edges:

\begin{equation}
\label{eq:cpm}
\scriptsize
\begin{aligned}
\textrm{Forces} \quad \quad F^t_j &:= f_F \left( F^{t-1}_j, \left\{O^{t-1}_i\right\}_{i|(i,j) \in J^{t-1}}\right) &\text{    s.t.    } J^{t-1} \sim  \rho^{t-1}_{\mathcal{O}},\\
\textrm{Objects} \quad \quad O^t_i &:= f_O \left( O^{t-1}_i, \left\{F^t_j\right\}_{j|(j,i)\in I^{t-1}}\right) &\text{    s.t.    } I^{t-1} \sim  \rho^{t-1}_{\mathcal{O}\leftrightarrow\mathcal{F}}.
\end{aligned}
\end{equation}
Thus, update functions $f_F$ and $f_O$ are force- and object-specific (respectively) and invariant to the number of inputs (i.e., size of the parent node set). When interventions occur at time step $\tilde{t}$, they introduce perturbations over object nodes, denoted as $\mathrm{do}(F^{\tilde{t}}_*,*\to \imath)$ representing externally applied action $F_*$ on object $O_\imath$ (e.g., hitting a billiard ball). Intuitively, this step captures how such external influences propagate forward in time, akin to resimulating the physical system from the intervention point onward: the model dynamically recomputes the subgraph for all subsequent time steps $t\geq \tilde{t}$ by resampling causal edges and updating node states via the controllers and transition functions, ensuring the causal graph reflects the altered dynamics (following Algorithm \ref{alg:intervention} in Appendix \ref{sec:algorithms}).

\subsubsection{Concrete Example: Two Colliding Balls}
To illustrate, consider a scenario with three balls, where Ball 1 collides with Ball 2 at time $t$ (Fig.~\ref{fig:example}a). The objects are represented as $\{O^t_1, O^t_2, O^t_3\}$, and a single force node $F^t_1$ mediates the interaction. Here, the interaction scope controller $\rho^t_{\mathcal{O}}$ assigns high probability to the edges $J^t = \left\{E\left(O^t_1,F^t_1\right), E\left(O^t_2,F^t_1\right)\right\}$, indicating that both Ball 1 and Ball 2 contribute to generating the collision force. Similarly, the effect attribution controller $\rho^t_{\mathcal{O}\leftrightarrow\mathcal{F}}$ assigns high probability to the edges $I^t = \left\{E\left(F^t_1,O^{t+1}_1\right), E\left(F^t_1,O^{t+1}_2\right)\right\}$, specifying that the force affects both balls post-collision. 
The resulting graph is illustrated in Fig.~\ref{fig:example}c.
In contrast, when the balls are far apart and no interaction occurs, both controllers would output empty edge sets, resulting in a sparse graph with only self-connections during that time period (e.g., Ball 3 in Fig.~\ref{fig:example}c).

\subsubsection{Inductive biases}
\label{sec:induc_bias}

To ground the flexible graph construction in realistic physical principles and mitigate the risk of overfitting to spurious connections, we incorporate two key inductive biases that reflect common patterns in object interactions. 
\textbf{Pairwise Interactions} restrict each force node to connect to exactly two different object nodes. 
This corresponds to a domain-motivated inductive bias that in many contact-like environments interactions are predominantly pairwise; it also keeps the controller’s action space tractable.
This restriction can be lifted later to generalize to hypergraphs for more complicated systems (e.g., 3-body problems). We model a \textbf{mirroring (consistency) constraint}: if a force is generated by a set of objects, it can only be attributed to (i.e., affect) objects within that set. This encourages physically plausible and interpretable graphs by preventing attribution to unrelated objects. Importantly, this constraint does not enforce equal or symmetric effects; the effect attribution controller can attribute a force to one or both participants (Sec.~\ref{model:controller}), and the object update $f_O$ can learn asymmetric influence magnitudes, including effectively zeroing influence on one side in domains with asymmetric interactions. Because $f_F$ and $f_O$ in Eq.~\ref{eq:cpm} are invariant to the number of parents, CPM’s update mechanism itself does not require pairwise interactions; extending the controllers to $
k$-ary selections is a natural generalization, although we do not evaluate it here. In physical collisions, forces come in equal and opposite pairs acting on both objects. This design ensures physical consistency while maintaining computational efficiency. In environments with asymmetric interactions (e.g., large objects unaffected by small ones), the learned weights in $f_O$ can effectively set the influence to zero (see Appendix \ref{appendix:inductive} for formal definitions). 

\section{Model}
\label{sec:Model}

We base our model implementation on the  
Contrastively-trained Structured World Model \citep[C-SWM;][]{DBLP:conf/iclr/KipfPW20}. 
The model consists of an \emph{object-centric vision encoder}, an \emph{action encoder}, and a \emph{transition function} (Fig. \ref{fig:edges-intervention-model}). We keep the structure of the vision and action encoders intact, but modify the transition function.

The \textit{vision encoder} is a CNN-based object extractor $E_\text{ext}$, operating directly on images and outputting $I$ feature maps. Each feature map $m^t_i = [E_\text{ext}(s^t)]_i$ acts as an object mask where $[\dots]_i$ is the selection of the $i^\text{th}$ feature map. An MLP-based object encoder $E_\text{enc}$ with shared weights across objects maps the flattened feature map $m^t_i$ to object latent representation: $O^t_i = E_\text{enc}(m^t_i)$. Additionally, an MLP-based \textit{action encoder} maps action $a^t$ to force latent representation: $F^t = A(a^t)$. Next, we introduce our new transition function (Sec.  \ref{model:transition}) before detailing how to construct the causal graph on the fly using reinforcement learning (Sec.  \ref{model:controller}).

\subsection{Causal Process Block}
\label{model:transition}

\begin{figure}[t]
    \centering
    \includegraphics[width=0.95\textwidth]{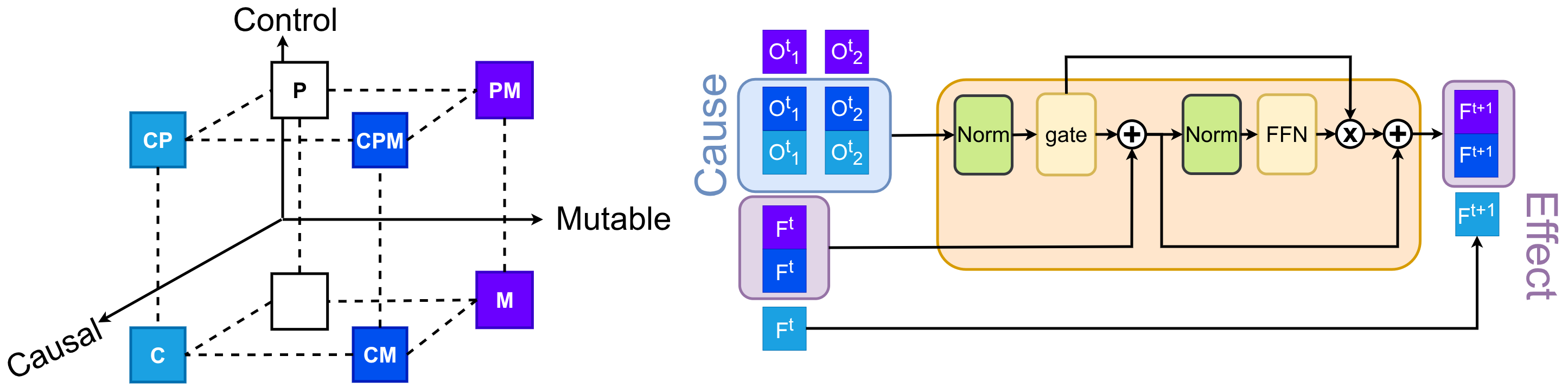}
    \vspace{-1.0em}
    \caption{\textbf{Causal Process Block} (illustrating $f_F$): Modified transformer with attention replaced by gate mechanism (see Appendix \ref{appendix:cpb}). Incoming causes $O^t_i$ are pre-selected by the causal controllers. Latent vectors are divided into causal (C), control (P), and mutable (M) regions, enforcing structured updates.}
    \label{fig:causal-process-net}
    \vspace{-1.0em}
\end{figure}

Our main innovation is the \textit{Causal Process Block} as a neural network implementation of a CPM (Fig. \ref{fig:causal-process-net}). Before introducing the technical details, we need to address a key challenge: not all components of force and object representations play the same role in causal interactions. 

\subsubsection{Structured Representations}
\label{model:struc_rep}
Let us revisit the example of collision between two balls (Fig.~\ref{fig:example}a). Their masses affect momentum transfer (causally relevant) but remain unchanged during the collision (immutable). In contrast, their velocities are both causally relevant and mutable. Meanwhile, visual properties like color may change due to lighting, but don't affect the collision dynamics (mutable but not causally relevant). To capture these distinctions and enable our model to learn interpretable encodings that naturally separate these different physical properties, we factorize our representations along three key binary subspaces of \textit{Causal Relevance} ($C$), \textit{Control Relevance} ($P$), and \textit{Mutability} ($M$). The binary nature of each subspace arises from selective routing within the CPM (Fig.~\ref{fig:causal-process-net}). 

Causal Relevance ($C$) describes whether a component influences the dynamics of other objects. For instance mass and velocity affect collision outcomes ($C=1$), but color does not ($C=0$).
Control Relevance ($P$) encodes whether a component is used by the control/policy functions for decision-making. For example, a controller deciding which balls are about to collide will rely on current positions and velocities ($P=1$), but will ignore other properties such as mass or purely visual features like color ($P=0$). Mutability ($M$) captures whether a component can change over time through interactions. For instance, an object being struck may change velocity ($M=1$), while its mass remains constant ($M=0$). Importantly, while the routing structure is fixed (based on C/P/M), the model learns end-to-end which features populate each subspace. We view this as an inductive bias that encourages interpretable factorization; it may be less suitable in domains where these axes are misaligned with the true generative factors, which we note as a limitation and a direction for future work.

\subsubsection{Technical Implementation}

We use two feedforward neural networks, $f_F(\dots ; \theta_F)$ and $f_O(\dots; \theta_O)$, shared by all the force and object nodes respectively. 
The force vector 
$F_j^t := \bigoplus_{\substack{C,P,M \in \{1,0\},  (C,P,M) \neq (0,0,0)}} F_j^{t, CPM} $ is the concatenation of all combinations of the $C$, $P$, $M$ dimensions except for $(C,P,M) = (0,0,0)$, which must be omitted since forces, by definition, do not contain subspaces that are irrelevant to the causal process.
This results in $2^3-1=7$ sub-vectors of equal size $d_F$, where a sub-vector's identity determines how it is processed by the neural networks: the object-update function $f_O$ and force-update function $f_F$ operate exclusively on the causally relevant subspace ($C=1$), while mutable parts are updated and immutable parts are copied unchanged ($M=0$; Fig. \ref{fig:causal-process-net}). 
The object vector $O^t_i$ is more straightforwardly divided into $2^3=8$ subvectors, i.e., all possible combinations of the $C$, $P$, $M$ dimensions, including the $(C,P,M) = (0,0,0)$ subspace: $O_i^t := \bigoplus_{\substack{C,P,M} \in \{1,0\}} O_i^{t, CPM}$.
The extra subspace is present here for the network to learn to shift visual input features that are irrelevant to the causal process into this subspace, for example the object's color in a collision event (See Appendices~\ref{appendix:vectors} and \ref{appendix:cpb} for details).
Note that the implementation of the causal process blocks $f_O\left(\dots\mid\theta_O \right)$ and $f_F\left(\dots\mid\theta_F \right)$ is similar to that of transformer blocks, but with the attention mechanism \citep{DBLP:conf/nips/VaswaniSPUJGKP17} replaced by indices of the chosen force and object nodes (tokens in transformers; see Appendix \ref{appendix:cpb} and Fig.~\ref{fig:causal-process-net} for more details). Unlike the attention mechanism of the transformer, $I^t,J^t$ can also be an empty set. This is analogous to transformer attention assigning zero weight to all the tokens, which they cannot do by design.

\subsection{Causal Controller}
\label{model:controller}

The main proposal of our model is that we perform graph construction through sequential decision making using the interaction scope and effect attribution controllers.
We treat causal discovery as a multi-agent RL problem. One agent (the interaction scope policy $\pi_{O}\left(\mathcal{O}^t\right) := \rho^t_{\mathcal{O}}$) determines the scope of interacting objects, and another (the effect attribution policy $\pi_{O\leftrightarrow F} \left(\mathcal{O}^t,\mathcal{F}^{t+1} \right) :=\rho^t_{\mathcal{O}\leftrightarrow\mathcal{F}}$) determines how force effects are attributed .

\begin{figure}[t]
    \centering
    \includegraphics[width=0.95\textwidth]{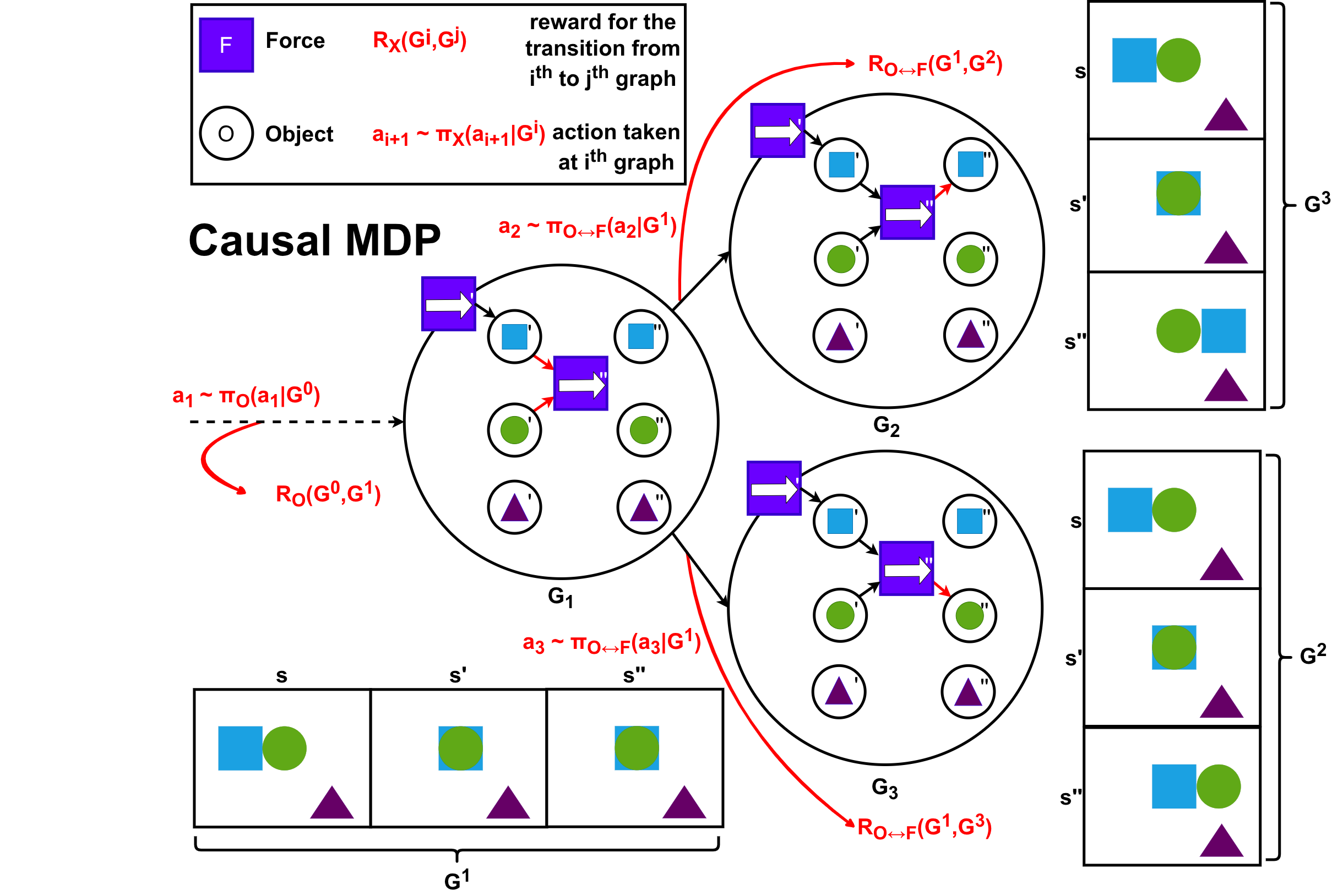}
    \vspace{-1.0em}
    \caption{\textbf{Causal MDP} used by the reactive agents to construct causal process graphs. Agents successively add edges to the causal graph. Each causal graph hypothesis corresponds to a potential sequence of frames.}
    \label{fig:tree}
    \vspace{-1.0em}
\end{figure}

More specifically, the chosen indices $I^t$ and $J^t$ are provided by the agents $\pi_{\mathcal{O}\leftrightarrow\mathcal{F}}$ and $\pi_{\mathcal{O}}$.
The two agents alternate outputting an action.
An action taken by $\pi_{\mathcal{O}}$ corresponds to two edge additions $E\left( O^t_i, F^{t+1} \right), E\left(O^t_j, F^{t+1}\right), i\neq j$ to the graph (selecting a pair of objects for interaction). 
Whereas an action taken by $\pi_{\mathcal{O}\leftrightarrow\mathcal{F}}$ results in either one or two edge additions $E\left( F^t,O^t_i \right), E\left( F^t,O^t_j \right)$ (attributing the force effect to either one or both objects; see Fig.~\ref{fig:tree}). 
The index set $I^t$ is sampled using the policy of the agent $\pi_{\mathcal{O}\leftrightarrow\mathcal{F}}$. Note that the policies only utilize the Control Relevant (P) features of the latent representations:
\begin{equation}
\scriptsize
\begin{aligned}
I^t \sim \pi_{\mathcal{O}\leftrightarrow\mathcal{F}}\left(  I^t\mid G^t,W_O,W_F \right) &= \operatorname{softmax}\left(Q_{\mathcal{O}\leftrightarrow\mathcal{F}}\left( G^t, I^t \mid W_O,W_F\right)\right)\\ &=\operatorname{softmax}\left( \frac{\left( F^{t,C1M} W_F \right)  \left(\left[ O_i^{t,C1M}; O_j^{t,C1M}; \left(O_i^{t,C1M} + O_j^{t,C1M}\right)/2 \right] W_O\right)^T} {d} \right),
\end{aligned}
\end{equation}

where $W_O\in\mathbb{R}^{4d_O\times d}$, $W_F\in\mathbb{R}^{4d_F\times d}$, and $Q_{\mathcal{O}\leftrightarrow\mathcal{F}}$ is the corresponding Q-value.
$J^t$, on the other hand, is sampled using the policy of the agent $\pi_{\mathcal{O}}$:

\begin{equation}
\scriptsize
\begin{aligned}
J^t &\sim \pi_{\mathcal{O}}\left(  J^t\mid G^t,W_{\tilde{O}} \right) = \sigma\left( Q_{\mathcal{O}}\left( G^t,J^t\mid G^t,W_{\tilde{O}} \right) \right) =  \sigma \left( \left( O_i^{t,C1M} W_{\tilde{O}} \right) \cdot \left(O_j^{t,C1M} W_{\tilde{O}} \right)  \right),
\end{aligned}
\end{equation}

where $W_{\tilde{O}}\in\mathbb{R}^{4d_O\times d}$, $\sigma$ is the sigmoid function, and $ Q_{\mathcal{O}}$ is the corresponding Q-value.

We then define separate reward functions for $\pi_{\mathcal{O}\leftrightarrow\mathcal{F}}$ and $\pi_{\mathcal{O}}$, modeled by MLPs parameterized by $\theta_{R_{\mathcal{O}\leftrightarrow\mathcal{F}}}$ and $\theta_{R_\mathcal{O}}$ respectively:

\begin{equation}
\label{eq:rewards}
\scriptsize
\begin{aligned}
    &R_{\mathcal{O}\leftrightarrow\mathcal{F}}\left(G^{t},G^{t+1} \mid \theta_{R_{\mathcal{O}\leftrightarrow\mathcal{F}}}\right) =
    \operatorname{MLP}\left(G^{t}_{V},\mathds{1}_{E\left( F^t,O^t_i \right)}, \mathds{1}_{E\left( F^t,O^t_j \right)}, G^{t+1}_{V} \middle| \theta_{R_{\mathcal{O}\leftrightarrow\mathcal{F}}} \right),\\
    &R_{\mathcal{O}}\left(G^{t},G^{t+1} \mid \theta_{R_\mathcal{O}} \right) =
    \operatorname{MLP}\left(G^{t}_{V},\mathds{1}_{E\left( O^t_i, F^{t+1} \right)\wedge E\left( O^t_j, F^{t+1} \right)}, G^{t+1}_{V} \middle| \theta_{R_\mathcal{O}} \right),
\end{aligned}
\end{equation}

where $G^{t} := \left( G^{t}_{V}, G^{t}_{E}\right)$ is the graph at time $t$ and $\mathds{1}_{E\left( \cdot,\cdot \right)}$ indicates the presence or absence of the edge $E\left( \cdot,\cdot \right)$. These reward functions are learned through inverse reinforcement learning \citep{DBLP:conf/icml/NgR00}, where the goal is to find a reward function whose corresponding optimal policy would select causal edges that would minimize prediction error.

\subsection{Training Procedure}

Edge selection and representation learning are tightly coupled: controllers require informative representations to choose edges, while learning informative representations depends on selecting reasonable edges. We therefore adopt a staged, EM-like alternating optimization \citep{dempster1977maximum} to stabilize learning (Stage 1 warm-starts representation/dynamics under fixed controllers; Stages 2–3 alternate reward learning and policy improvement).  We choose discrete, all-or-nothing edges to preserve sparse computation and avoid reverting to dense message passing during training. The overall goal is to learn a predictive world model (CPM) whose structure is determined by the policies ($\pi_{\mathcal{O}}, \pi_{\mathcal{O}\leftrightarrow\mathcal{F}}$). This requires optimizing both the model parameters $\Theta$ and the policy parameters $\Psi:=\left[W_O,W_F,W_{\tilde{O}}\right]$.

\textbf{1. Prediction.}
At this stage, we freeze all the weights except $
\Theta = \left[\theta_V;\theta_A;\theta_O;\theta_F\right]$ and sample edges $\left\{I^\tau,J^\tau\right\}_{\tau}$ using frozen controllers. This allows the vision encoder, action encoder, and transition functions ($f_O, f_F$) to learn useful representations before the controllers get the chance to optimize their behavior on these representations. We train the model parameters $\Theta$ using contrastive loss \citep{DBLP:conf/iclr/KipfPW20}:

\begin{equation}
\scriptsize
\label{eq:contrastiveLoss}
\begin{aligned}
    \mathcal{L}_\text{pred}\left( \Theta | \beta, \mathcal{D}_\text{pred}\right) &= \left\|\operatorname{CPM}\left(\operatorname{V}\left(s^t\middle| \theta_V\right),\operatorname{A}\left(a^t\middle| \theta_A\right),\left(I^\tau,J^\tau\right)_{t_\tau} \middle| \theta_O,\theta_F \right)- \operatorname{V}\left(s^{t+1} \middle| \theta_V \right) \right\| \\&+ \operatorname{max}\left( 0,\beta - \left\| \operatorname{V}\left(\tilde{s}^t\middle| \theta_V \right)- \operatorname{V}\left(s^{t+1}\middle| \theta_V \right) \right\| \right)
\end{aligned}
\end{equation}

where $\mathcal{D}_\text{pred}=\left\{ \left(s^t, a^t, s^{t+1}\right), \tilde{s}^t, \left\{\left(G^{t_\tau}, I^{t_\tau},J^{t_\tau},G^{t_{\tau+1}}\right)\right\}_{\tau}
\right\}_t$, $\operatorname{V}$ and $\operatorname{A}$ are the vision and action encoders, $\tilde{s}^t$ is a negative example sampled from the experience buffer, and the hinge margin $\beta$ is set to 1 \citep[following][]{DBLP:conf/iclr/KipfPW20}.

\textbf{2. Expectation.}
In the second stage, we freeze all the weights but $\Theta_R= \left[ \theta_{R_{\mathcal{O}\leftrightarrow\mathcal{F}}};\theta_{R_{\mathcal{O}}} \right]$ and use temporal difference (TD) loss \citep{DBLP:journals/ml/WatkinsD92} to learn reward functions:
\begin{equation}
\scriptsize
\label{eq:rewardLoss}
\begin{aligned}
    \mathcal{L}_\text{TD}\left( \Theta_R \mid \mathcal{D}_\text{TD},\Psi \right) 
    &= \sum_{X\in\{ {\mathcal{O}\leftrightarrow\mathcal{F}},\mathcal{O}\} } \sum_\tau \left\| 
    \underbrace{R_X\!\left(G^\tau,G^{\tau+1}\middle|\theta_{R_X}\right)}_{\text{learned}}
    \;-\;
    \underbrace{\left( 
    Q_X\left( G^\tau, I^\tau_X \mid \Psi \right)
    -\gamma \underset{I^{\tau+1}_X}{\max} \, Q_X\left( G^{\tau+1}, I^{\tau+1}_X \mid \Psi \right) 
    \right)}_{\text{target}} \right\|
\end{aligned}
\end{equation}

where $\mathcal{D}_\text{TD} = \left\{\left(G^\tau, I^\tau_\mathcal{O},I^\tau_{\mathcal{O}\leftrightarrow\mathcal{F}},G^{\tau+1}\right)\right\}_{\tau}$ and we set $\gamma=0.9$.

\textbf{3. Maximization.} During the third stage, we freeze all but the policy parameters $\Psi$ and use the same TD loss from above with target and learned terms reversed. The agents learn to select edges based on the rewards $R_{\mathcal{O}\leftrightarrow\mathcal{F}}$ and $R_{\mathcal{O}}$.

The overall optimization landscape is complex (see Appendix \ref{appendix:optim} for further optimization details); the optimum represents a state where the agents select the sparse causal graph that yields the minimal prediction loss for the CPM, and simultaneously, the reward MLPs stabilize under the IRL objective.

\section{Experiments}
\label{sec:expriments}

\begin{figure}[t]
    \begin{center}
        \centerline{\includegraphics[width=0.95\textwidth]{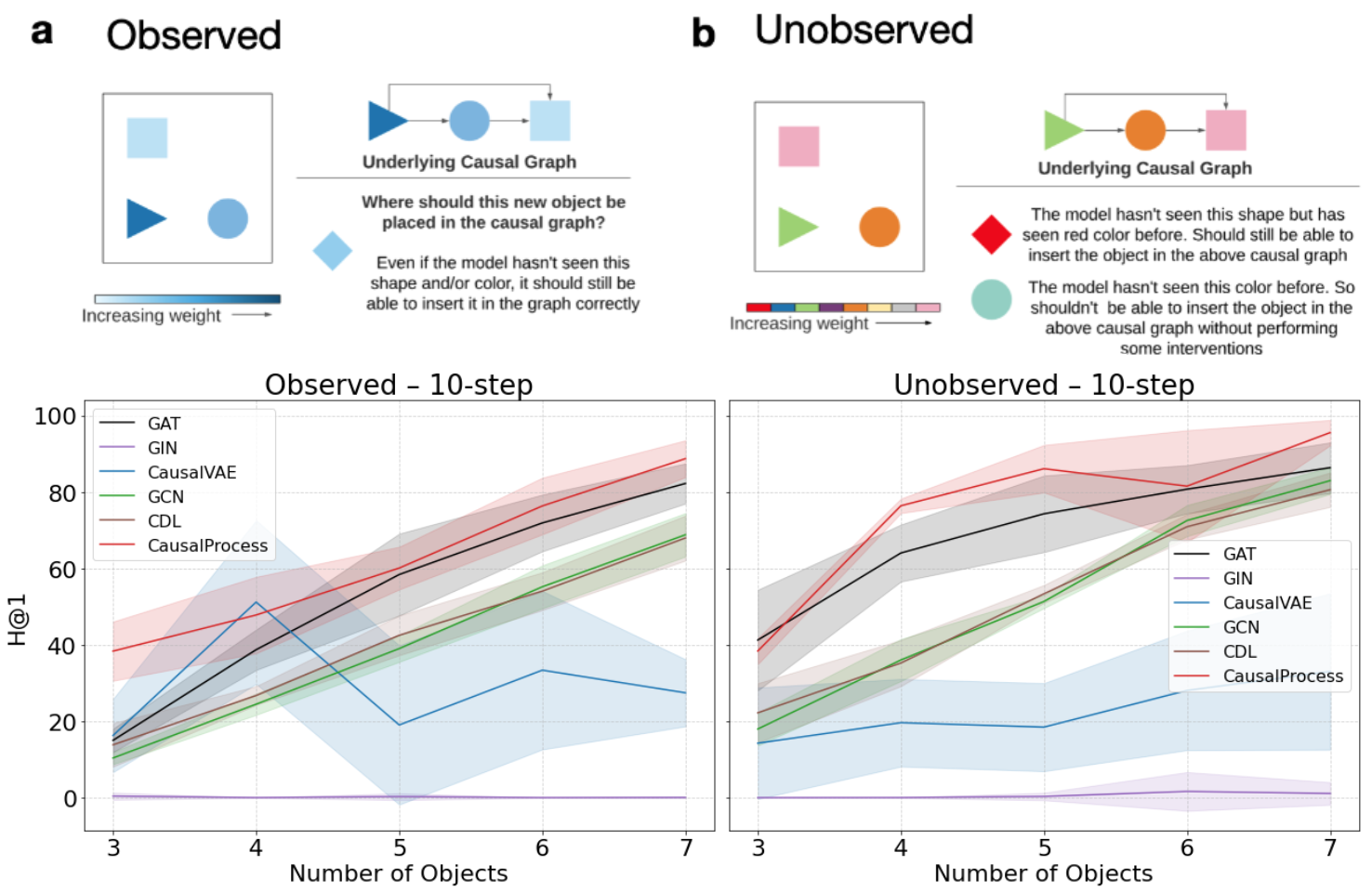}}
    \end{center}
    \vspace{-1.5em}
    \caption{\textbf{Prediction results for a synthetic physics environment} in a) observed and b) unobserved settings \citep{DBLP:conf/nips/KeDMGLBRMBP21}. \textbf{Top:} Description of the task. \textbf{Bottom:} Prediction metric vs number of objects after 10 steps (average of 10 seeds). For remaining baselines, see Fig.~\ref{fig:h@1-10-step-other-baselines} in Appendix \ref{sec:plots}.}
    \label{fig:env}
    \vspace{-1em}
\end{figure}

We hypothesize that our model outperforms models that assume dense causal graphs to capture physical interactions in: 1) longer prediction horizons; 2) test-time generalization across unobservable properties; 3) robustness with regards to the number of objects in the scene; 4) solving downstream tasks. We use the \textit{physics environment} designed by \citet{DBLP:conf/nips/KeDMGLBRMBP21} to empirically answer these questions (Fig.~\ref{fig:env} top). The environment consists of different objects colored according to their weights. The only force in this environment is pushing (double-pushes are not allowed) and only heavier objects can push lighter ones. The environment has two settings: an \textit{observed} setting (Fig.~\ref{fig:env}a) where weight corresponds to the intensity of a particular color and an \textit{unobserved} setting (Fig.~\ref{fig:env}b) where different colors do not systematically map to different weights.

\paragraph{Benchmark choice and empirical scope.}
We adopt the synthetic physics benchmark of \citet{DBLP:conf/nips/KeDMGLBRMBP21} because it is an established
testbed for visual causal discovery in model-based RL and enables direct comparison to a broad set
of prior baselines under a standardized protocol.
The benchmark is intentionally controlled: interactions are sparse and pairwise, observations are
strongly object-centric, and the only explicit interaction is a pushing force (with additional constraints
such as no double-pushes and asymmetric pushing between heavy/light objects).
These restrictions allow us to isolate the main claim of CPMs, that dynamically constructing
\emph{sparse, time-varying} causal graphs improves long-horizon prediction and downstream planning, while
keeping confounds minimal. We view evaluation in more diverse physical domains as an important
next step (see Sec.~\ref{discussion}).

\subsection{Comparison Baselines}
\label{subsec:baselines}

We compare our model against 5 causal and graph neural network baselines: a graph attention network (GAT) \citep{velickovic2018graph}, a graph isomorphism network (GIN) \citep{xu2018how}, a causal variatiational auto-encoder (CausalVAE) \citep{9578520}, a graph convolutional network (GCN) \citep{kipf2017semisupervised}, and a causal dynamics learning network (CDL) \citep{DBLP:journals/corr/abs-2206-13452}. Additionally, we compare the model against the 5 baselines from \citet{DBLP:conf/nips/KeDMGLBRMBP21}: a graph neural network (GNN) \citep{DBLP:journals/tnn/ScarselliGTHM09}, a transformer network \citep{DBLP:conf/nips/VaswaniSPUJGKP17}, a recurrent independent mechanisms (RIM) network \citep{DBLP:conf/iclr/GoyalLHSLBS21}, a schema / object-file factorization network (SCOFF) \citep{goyal2021factorizing}, and a modular network that has a separate MLP to model each object's dynamics. We do not include RL-based tabular causal discovery methods as experimental baselines because they output a single global graph over pre-defined variables and do not define a visual world-model predictor under our evaluation metrics \citep{DBLP:conf/iclr/ZhuNC20, DBLP:conf/atal/SauterBAP24, DBLP:journals/tmlr/Duong0HN25}.

\subsection{Prediction Metrics}
\label{subsec:prediction}

To investigate robustness towards the length of prediction horizons, we trained the model to make 1-step predictions in the \textit{Observed} setting with several objects and then tested for 5 (Fig.~\ref{fig:h@1-1-5-steps}) and 10 steps (Fig.~\ref{fig:env}a bottom, Fig.~\ref{fig:h@1-10-step-other-baselines}). We used Hits at Rank 1 (H$@$1) to measure model performance as an all-or-nothing metric measuring how often the rank of the predicted representation was 1 when ranked against all reference state representations. Here, our model broadly outperformed the baseline models, with the gap increasing over longer time horizons.

Next, to estimate the test-time generalization across unobservable properties, we trained our model in the \textit{Unobserved} setting where generalization at test time is harder due to previously unseen colors (weights). Again, our model broadly outperformed the baselines displaying capacity to generalize also in this domain (Fig.~\ref{fig:env}b bottom; see Fig.~\ref{fig:h@1-10-step-other-baselines} and Fig.~\ref{fig:h@1-1-5-steps} for more results).

\begin{figure}[t]
    \centering
    \includegraphics[width=0.95\textwidth]{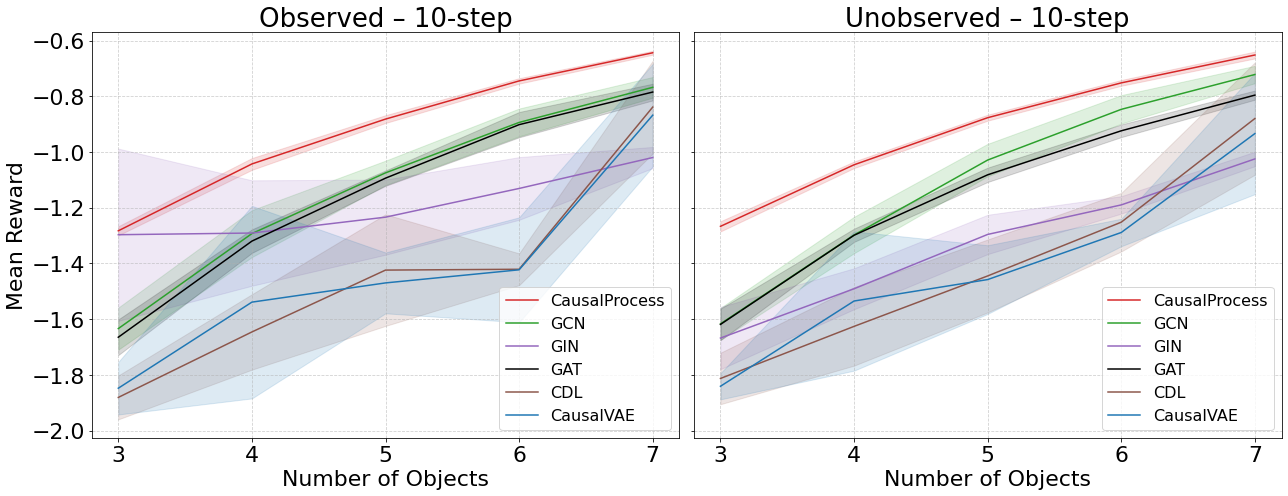}
    \vspace{-1.0em}
    \caption{\textbf{Downstream RL results} over number of objects. Mean reward vs number of objects. All results are the average of 10 seeds. For remaining baselines, see Fig.~\ref{fig:mean-reward-other-basleines} in Appendix \ref{sec:plots}.}
    \label{fig:mean_reward}
    \vspace{-1.5em}
\end{figure}

\subsection{Downstream RL tasks}
\label{subsec:ablations_graph_recovery}

To make sure the above metrics overlap with the learned model's usefulness for downstream tasks, we also tested our CPM's capacity to serve as a world model for a model-based RL agent. The agent's task was to move an object to a certain location each taken step resulting in negative reward. In both Observed and Unobserved settings, the agent with CPM as model of the environment broadly outperformed the baselines for all objects in 10-step unrolling of the learned model (Fig.~\ref{fig:mean_reward}, Fig. ~\ref{fig:mean-reward-other-basleines}).

\subsection{Ablations and Graph Recovery}
\label{subsec:downstream_rl}

We include a component ablation study in Appendix~\ref{app:ablation}. We additionally evaluate (i) graph recovery against the simulator's ground-truth interaction graph
and (ii) the semantics of the learned $(C,P,M)$ subspaces via linear probing; see
Appendix~\ref{app:graph_and_repr}.

\section{Discussion}
\label{discussion}

 In this paper, we introduced the Causal Process Framework (CPF) as a novel approach for modeling the dynamics of physical object interactions. Our key contribution is the Causal Process Model (CPM), which implements this framework by treating the edge distributions inherent to CPF as a reinforcement learning policy. Instead of the soft, dense connections typical of many baselines \citep{velickovic2018graph, goyal2021factorizing, xu2018how, 9578520, kipf2017semisupervised, DBLP:journals/corr/abs-2206-13452, DBLP:journals/tnn/ScarselliGTHM09, DBLP:conf/nips/VaswaniSPUJGKP17, DBLP:conf/iclr/GoyalLHSLBS21}, our model employs RL agents to dynamically construct sparse, time-varying causal graphs. Our experiments in a simulated physics environment \citep{DBLP:conf/nips/KeDMGLBRMBP21} show that this approach not only improves prediction accuracy and downstream task performance compared to baselines, but also excels in generalization and scalability.

The superior performance of our model, particularly over longer prediction horizons and with a varying number of objects, lends strong support to our central hypothesis. We argue that by explicitly modeling only active causal links, the CPM avoids the pitfalls of dense message-passing architectures \citep{DBLP:journals/corr/abs-2406-04267, DBLP:conf/iclr/0002Y21, DBLP:conf/iclr/BarberoVSBG24, DBLP:conf/icml/GiovanniGBLLB23, DBLP:journals/tmlr/GiovanniRBDLMV24, DBLP:conf/iclr/ToppingGC0B22, DBLP:journals/tnn/ScarselliGTHM09, DBLP:journals/corr/battaglia_gnn}. Our discrete, ``all-or-nothing'' connections, determined by a goal-oriented RL agent, preserve the salience of individual interactions. This leads to more robust and precise world models, which proved crucial for the model-based RL agent's success in downstream tasks. Furthermore, the model's ability to generalize to unobserved object properties suggests that it learns an underlying model of physical dynamics rather than memorizing superficial correlations.

Despite these promising results, the present work has several limitations that open clear avenues for future research. A primary limitation is that our empirical evaluation is conducted in a single benchmark environment with strongly object-centric visuals, pairwise interactions, and a single force type (pushing). While this setting provides a clean comparison against established baselines, it does not by itself demonstrate robustness to substantially more complex physical interactions or perception. Importantly, these restrictions are not fundamental to the Causal Process Framework / CPM.  Our current pairwise interaction constraint is a deliberate inductive bias (Sec.~\ref{sec:induc_bias}) and can be lifted to support hypergraph-style interactions (e.g., three-body effects) by allowing force nodes to connect to sets of objects rather than pairs.  The object-centric structure in our experiments reflects the chosen vision front-end (Sec.~\ref{sec:Model}) rather than a requirement of CPMs. An important direction is to pair CPMs with more general perception modules that handle clutter, occlusion, and imperfect segmentation. Additionally, we do not empirically compare against continuous edge-selection relaxations (e.g., Concrete/Gumbel masking) \citep{DBLP:conf/iclr/JangGP17,DBLP:conf/iclr/LouizosWK18,DBLP:conf/iclr/MaddisonMT17} and we do not provide an extensive hyperparameter sensitivity analysis; we leave these evaluations to future work.

\acks{Funded by the Deutsche Forschungsgemeinschaft (DFG, German Research Foundation) under Germany’s Excellence Strategy – EXC number 2064/1 – Project number 390727645. C.M.W is supported by the European Research Council (ERC) under the European Union’s Horizon 2020 research and innovation programme (C4: 101164709), the Deutsche
Forschungsgemeinschaft (German Research Foundation, DFG) under
Germany’s Excellence Strategy (EXC 3066/1 ``The Adaptive Mind'',
Project No. 533717223), and the Excellence Cluster ``Reasonable AI'' by the Deutsche Forschungsgemeinschaft (German Research Foundation, DFG) under Germany’s Excellence Strategy – EXC-3057. The authors thank the International Max Planck Research School for Intelligent Systems (IMPRS-IS) for supporting Turan Orujlu.} The authors also thank Antonio Orvieto, Siyuan Guo, and Patrik Reizinger for insightful discussions.

\bibliography{main, chris}

\appendix

\section{Object-centric Causal Dynamics}
\label{sec:object-centric_causal_dynamics}
Consider two objects $\textcolor{magentaObj}{O_1}$ and $\textcolor{magentaObj}{O_2}$ (depicted in \textcolor{magentaObj}{magenta}) with a force $\textcolor{violetFor}{F}$ (depicted in \textcolor{violetFor}{violet}) acting on them:

\begin{align*}
&\textcolor{magentaObj}{O_1}\textcolor{violetFor}{\xrightarrow{F}}\textcolor{magentaObj}{O_2}
\end{align*}

This recovers the familiar structure of a directed acyclic graphs (DAGs) from Pearl's causal formalism \cite{Pearl2009CausalII}. However, in physical interactions, such as in a collision, it is not always clear which object is the ``cause'' since both are affected simultaneously. A more intuitive representation would be a bidirectional edge:
\begin{align*}
\textcolor{magentaObj}{O_1}\textcolor{violetFor}{\xleftrightarrow{F}}\textcolor{magentaObj}{O_2}
\end{align*}

However, DAGs prohibit cycles and bidirectional edges. To resolve this, we introduce \textit{temporal dynamics} which represent causal effects as unfolding over time rather than as a simultaneous influence. Thus, a collision between object $\textcolor{magentaObj}{O_1}$ and object $\textcolor{magentaObj}{O_2}$ yields forces $\textcolor{violetFor}{F_2}$ and $\textcolor{violetFor}{F_3}$ as emerging from the past state and influencing future object states: 
\begin{center}
\begin{tikzcd}[column sep=6em, row sep=.7em] 
  {\color{magentaObj} O_1^t}
    \arrow[r,
      "\textcolor{violetFor}{F_1}"{pos=0.5}, 
      draw=violetFor]
    \arrow[dr,
      "\textcolor{violetFor}{F_2}"{pos=0.6, xshift=-10pt, yshift=2pt}, 
      draw=violetFor]
  &
  {\color{magentaObj} O_1^{t+1}}
  \\
  {\color{magentaObj} O_2^t}
    \arrow[r, swap,
      "\textcolor{violetFor}{F_4}"{pos=0.5}, 
      draw=violetFor]
    \arrow[ur, swap,
      "\textcolor{violetFor}{F_3}"{pos=0.6, xshift=-10pt, yshift=-2pt}, 
      draw=violetFor]
  &
  {\color{magentaObj} O_2^{t+1}}
\end{tikzcd}
\end{center}

Yet, this representation still has drawbacks. Specifically, we break the identity of the force $\textcolor{violetFor}{F}$ into $\textcolor{violetFor}{F_2}$ and $\textcolor{violetFor}{F_3}$ which, in principle, can act as separate causal links ($\textcolor{violetFor}{F_1}$ and $\textcolor{violetFor}{F_4}$ can be thought of as inertia).
This becomes apparent when interventions are applied. Let us imagine that somebody picks up object $\textcolor{magentaObj}{O_1}$ just before it collides with $\textcolor{magentaObj}{O_2}$. This can be represented by a do-calculus-like intervention applied to either $\textcolor{magentaObj}{O_1^t}$ or $\textcolor{magentaObj}{O_1^{t+1}}$:

\begin{center}
\begin{minipage}{0.45\textwidth}
\begin{tikzcd}[column sep=6em, row sep=.7em] 
  {\raisebox{0.5ex}{\scalebox{0.8}{\faHammer}}{\color{magentaObj} O_1^t}}
    \arrow[r,
      "\textcolor{violetFor}{F_1}"{pos=0.5}, 
      draw=violetFor]
    \arrow[dr,
      "\textcolor{violetFor}{F_2}"{pos=0.6, xshift=-10pt, yshift=2pt}, 
      draw=violetFor]
  &
  {\color{magentaObj} O_1^{t+1}}
  \\
  {\color{magentaObj} O_2^t}
    \arrow[r, swap,
      "\textcolor{violetFor}{F_4}"{pos=0.5}, 
      draw=violetFor]
    \arrow[ur, swap,
      "\textcolor{violetFor}{F_3}"{pos=0.6, xshift=-10pt, yshift=-2pt}, 
      draw=violetFor]
  &
  {\color{magentaObj} O_2^{t+1}}
\end{tikzcd}
\end{minipage}
\quad  
\begin{minipage}{0.45\textwidth}
\begin{tikzcd}[column sep=6em, row sep=.7em]
  {\color{magentaObj} O_1^t}
    \arrow[dr,
      "\textcolor{violetFor}{F_2}"{pos=0.5, xshift=-10pt, yshift=2pt}, 
      draw=violetFor]
  &
  {\raisebox{0.5ex}{\scalebox{0.8}{\faHammer}}{\color{magentaObj} O_1^{t+1}}}
  \\
  {\color{magentaObj} O_2^t}
    \arrow[r, swap,
      "\textcolor{violetFor}{F_4}"{pos=0.5}, 
      draw=violetFor]
  &
  {\color{magentaObj} O_2^{t+1}}
\end{tikzcd}
\end{minipage}
\end{center}

When the intervention is applied to $\textcolor{magentaObj}{O_1^t}$, the graph structure is preserved, thus implying a no-collision scenario (one of the balls was lifted). Yet, the same graph can also imply a collision scenario. 
This kind of setup necessitates having causal links between objects that can potentially collide irrespective of the actualization of said collision. 
While this approach can work in principle, it results in extremely dense graphs with complete subgraphs per time step, especially in cluttered scenes. Ideally, we would like to have causal links in our graph if there is an actualized interaction between the involved objects.

On the other hand, intervening on $\textcolor{magentaObj}{O_1^{t+1}}$ results in a graph a with counter-intuitive interpretation: $\textcolor{magentaObj}{O_1}$ gets lifted at time step $t+1$, while $\textcolor{magentaObj}{O_2}$ behaves as if a collision has happened. This is due to the split of $\textcolor{violetFor}{F}$ into $\textcolor{violetFor}{F_2}$ and $\textcolor{violetFor}{F_3}$ since an intervention removes $\textcolor{violetFor}{F_3}$ while leaving $\textcolor{violetFor}{F_2}$ untouched. To tackle the aforementioned issues, let us re-imagine force edges as nodes and re-introduce $\textcolor{violetFor}{F_2}$ and $\textcolor{violetFor}{F_3}$ as a single node $\textcolor{violetFor}{F}$ and extend the time horizon by a step.

\begin{center}
\begin{tikzcd}[column sep=6em, row sep=0em]
  {\color{magentaObj} O_1^{t-1}}
    \arrow[rr, draw=violetFor, shorten <=2pt, shorten >=2pt] 
  & & 
  {\color{magentaObj} O_1^t}
    \arrow[rr, draw=violetFor, shorten <=2pt, shorten >=2pt] 
    \arrow[dr, draw=violetFor, shorten <=2pt, shorten >=2pt] 
  & & 
  {\color{magentaObj} O_1^{t+1}} \\ 

  & & & 
  {\color{violetFor} F}
    \arrow[ur, draw=violetFor, shorten <=2pt, shorten >=2pt] 
    \arrow[dr, draw=violetFor, shorten <=2pt, shorten >=2pt] 
  & \\ 

  {\color{magentaObj} O_2^{t-1}}
    \arrow[rr, draw=violetFor, shorten <=2pt, shorten >=2pt] 
  & & 
  {\color{magentaObj} O_2^t}
    \arrow[rr, draw=violetFor, shorten <=2pt, shorten >=2pt] 
    \arrow[ur, draw=violetFor, shorten <=2pt, shorten >=2pt] 
  & & 
  {\color{magentaObj} O_2^{t+1}} 
\end{tikzcd}
\end{center}

Now, imagine, just like before, the ball $\textcolor{magentaObj}{O_1}$ gets picked up at time step $t$. In do-calculus terms, this amounts to intervention to $\textcolor{magentaObj}{O_1^t}$ which results in mutilation of the edge $\textcolor{magentaObj}{O_1^{t-1}} \textcolor{violetFor}{\rightarrow} \textcolor{magentaObj}{O_1^t}$

\begin{center}
\begin{tikzcd}[column sep=6em, row sep=0em]
  {\color{magentaObj} O_1^{t-1}}
  & & 
  {\raisebox{0.5ex}{\scalebox{0.8}{\faHammer}}\,{\color{magentaObj}O_1^t}} 
    \arrow[rr, draw=violetFor, shorten <=2pt, shorten >=2pt] 
    \arrow[dr, draw=violetFor, shorten <=2pt, shorten >=2pt] 
  & & 
  {\color{magentaObj} O_1^{t+1}} \\ 

  & & & 
  {\color{violetFor} F}
    \arrow[ur, draw=violetFor, shorten <=2pt, shorten >=2pt] 
    \arrow[dr, draw=violetFor, shorten <=2pt, shorten >=2pt] 
  & \\ 

  {\color{magentaObj} O_2^{t-1}}
    \arrow[rr, draw=violetFor, shorten <=2pt, shorten >=2pt] 
  & & 
  {\color{magentaObj} O_2^t}
    \arrow[rr, draw=violetFor, shorten <=2pt, shorten >=2pt] 
    \arrow[ur, draw=violetFor, shorten <=2pt, shorten >=2pt] 
  & & 
  {\color{magentaObj} O_2^{t+1}} 
\end{tikzcd}
\end{center}

While the problem of splitting of the force identity seems to be resolved here, the graph structure modeling the collision remains preserved despite the intervention. 
As mentioned before, this can be addressed by complete subgraphs per time step, which is not desirable for our purposes.
This problem arises due to the inclusion of time dynamics into our graphs. Unlike in Pearlian Causality, in physics, interventions at a time step have implications for the causal connections corresponding to downstream time steps. To account for that, we have to re-imagine interventions under a new framework that takes physical processes and time into account (see Algorithm \ref{alg:intervention}).

\section{Algorithms}
\label{sec:algorithms}

\begin{algorithm2e}[H]
\caption{Interventions under Causal Process Framework}
\label{alg:intervention}

\KwIn{$\mathcal{F}^t, \mathcal{O}^t, f_{F}, f_{O}, \rho^t_{{\mathcal{O}}\leftrightarrow\mathcal{F}}, \rho^t_{\mathcal{O}}, \mathrm{do}\left(F^{\tilde{t}}_*,*\to\imath\right), G_{\mathcal{O}^1:\mathcal{O}^T}, t\in \{1,\dots,T\}, i \in \{1,\dots,I\}, j \in \{1,\dots,J\}$}
\KwOut{$\widetilde{G}_{\mathcal{O}^1:\mathcal{O}^T}$}

$\widetilde{\mathcal{O}}^{\tilde{t}-1}:= \mathcal{O}^{\tilde{t}-1}$\;
$\widetilde{\mathcal{F}}^{\tilde{t}} := \left\{F^{\tilde{t}}_*\right\} \dot{\cup} \mathcal{F}^{\tilde{t}}$\;
$J^{\tilde{t}} := \left\{(i,j)\right\}$ s.t.\ $E\left(O^{\tilde{t}-1}_i,F^{\tilde{t}}_j \right)\in G^{\mathcal{E}}_{\mathcal{O}^{\tilde{t}-1}:\mathcal{F}^{\tilde{t}}}$\;
$\widetilde{G}_{\mathcal{O}^1:\mathcal{O}^{\tilde{t}-1}} := G_{\mathcal{O}^1:\mathcal{O}^{\tilde{t}-1}}$\;

\For{$t = \tilde{t}$ \KwTo $T$}{
    $\widetilde{G}_{\mathcal{O}^1:\mathcal{F}^{t}} := \left(\widetilde{\mathcal{F}}^t \dot{\cup} \widetilde{G}^{\mathcal{V}}_{\mathcal{O}^1:\mathcal{O}^{t-1}}, \left\{ E\left(\widetilde{O}^{t-1}_i, \widetilde{F}^t_j\right) \right\}_{(i,j)\in J^t} \dot{\cup} \widetilde{G}^{\mathcal{E}}_{\mathcal{O}^1:\mathcal{O}^{t-1}} \right)$ \tcp*{graph}
    $I^t \sim \rho^t_{\widetilde{\mathcal{O}}\leftrightarrow \widetilde{\mathcal{F}}}$ \tcp*{for->obj edges}

    \uIf{$t=\tilde{t}$}{
        $I^t:= \{(*,\imath)\} \cup \left(I^t \setminus \{ (\cdot,\imath) \} \right)$ \tcp*{intervention}
    }
    
    \For{$i = 1$ \KwTo $I$}{
        $\widetilde{O}^t_i := f_{{O}}\left( \widetilde{O}^{t-1}_i, \left\{ \widetilde{F}^t_j \right\}_{j|(j,i)\in I^t} \right)$ \tcp*{object nodes}
    }

    $\widetilde{G}_{\mathcal{O}^1:\mathcal{O}^{t}} := \left(\widetilde{\mathcal{O}}^t \dot{\cup} \widetilde{G}^{\mathcal{V}}_{\mathcal{O}^1:\mathcal{F}^{t}}, \left\{ E\left(\widetilde{F}^t_j, \widetilde{O}^t_i\right) \right\}_{(j,i)\in I^t} \dot{\cup} \widetilde{G}^{\mathcal{E}}_{\mathcal{O}^1:\mathcal{F}^t} \right)$ \tcp*{graph}
    $J^t \sim \rho^t_{\widetilde{\mathcal{O}}}$ \tcp*{obj->for edges}

    \For{$j = 1$ \KwTo $J$}{
        $\widetilde{F}^{t+1}_j := f_{{F}}\left( \widetilde{F}^t_j, \left\{ \widetilde{O}^t_i \right\}_{i|(i,j)\in J^t} \right)$ \tcp*{force nodes}
    }
}
\Return{$\widetilde{G}_{\mathcal{O}^1:\mathcal{O}^T}$}
\end{algorithm2e}

\section{Model details}

\subsection{Inductive bias}
\label{appendix:inductive}
We introduce two inductive biases: (1) limiting each force node to interact with exactly two objects to reflect pairwise interactions, and (2) enforcing a bidirectional mirroring constraint to ensure temporal coherence in causal attribution.
Formally the latter is defined as:

\begin{align*}
\forall i,j,k,t: &\left\{ E(O^t_i, F^{t+1}_j), E(O^t_k, F^{t+1}_j) \right\} \subset G^t_E \implies \\
&E(F^{t+1}_j, O^{t+1}_i) \in G^t_E \vee E(F^{t+1}_j, O^{t+1}_k) \in G^t_E,\\
\forall i,j,t: &E(F^{t+1}_j, O^{t+1}_i) \in G^t_E \implies E(O^t_i, F^{t+1}_j) \in G^t_E.
\label{ass:mirroring}
\end{align*}

\subsection{Vector constraints}
\label{appendix:vectors}

\textit{Causal} relevance is coded by $C$. Perturbing the sub-vectors of the parent with $C=0$ does not affect the child nodes, i.e., only the causally-relevant $C=1$ sub-vectors affect the child nodes: 
\begin{equation*}
\scriptsize
\begin{aligned}
    &\forall t,i: 
    F^{t+1,1PM}_j = \widetilde{F}_j^{t+1,1PM} \implies  f_O\left(O^t_i, \left\{ F^{t+1}_j \right\}_{j\mid (j,i)\in I^t}; \theta_O \right) = f_O\left(O^t_i, \left\{ \widetilde{F}^{t+1}_j \right\}_{j\mid (j,i)\in I^t}; \theta_O \right).
\end{aligned}
\end{equation*}

\textit{Control} relevance is coded by $P$. Two force vectors whose sub-vectors with $P=1$ are identical have identical control functions that are conditioned on them, i.e., control functions are conditioned only on the control-relevant sub-vectors :

\begin{align*}
    &\forall t:  
    F_j^{t+1,C1M} = {\widetilde{F}_j}^{t+1,C1M} \implies \rho^t_{\mathcal{O}\leftrightarrow\widetilde{\mathcal{F}}} = \rho^t_{\mathcal{O}\leftrightarrow\mathcal{F}}.
\end{align*} 

Lastly, \textit{mutability} is coded by $M$. If $M=0$, the corresponding sub-vector does not change over time, i.e., an immutable sub-vector does not change over time: $\forall t,j:  F_j^{t,CP0}= F_j^{t+1,CP0}$. 

\subsection{Causal Process Block}
\label{appendix:cpb}
Given the data $F^t$, $O^t_1$, \dots, $O^t_n$, and the chosen indices $J^t$, we calculate $F^{t+1}$ in the following way:

\begin{align*}
    F^{t+1} := f_F\left(F^t,O^t_1,\dots,O^t_n,J^t\mid\theta_F\right),
\end{align*}

with $O^{t+1}$ also calculated similarly.
\begin{equation*}
\begin{aligned}
    &\text{gate} :=  \frac{1}{|J^t|}\sum_{i\in J^t} \left(O^{t,1PM}_i W^F_{\text{gate}}\right) W^F_\text{output},\\
    &\text{residual} := \text{gate} + F^{t,CP1},\\
    &F^{t+1,CP1} := \chi_{\text{gate} \neq 0}\odot\operatorname{FFN}\left(\operatorname{Norm}\left(\text{residual}\right)\right) + \text{residual},\\
    &F^{t+1} := \left[F^{t+1,CP1}; F^{t,CP0} \right],
\end{aligned}
\end{equation*}
where $\theta_F:=\left\{ W^F_{\text{gate}}, W^F_{\text{output}}, W^F_1, W^F_2, b^F_1, b^F_2 \right\}$ $\operatorname{FFN}$ is a feed-forward neural network $\operatorname{FFN}(x) := \operatorname{max}\left( 0,xW^F_1+b^F_1\right)W_2^F+b^F_2$, $W^F_{\text{gate}}$ is the analogue of the attention mechanisms value token projection, and $W^\mathcal{F}_{\text{output}}$ is again the analogous out-projection that maps the token from the attention dimension back to residual dimension \citep{DBLP:conf/nips/VaswaniSPUJGKP17}

\section{Optimization details}
\label{appendix:optim}

During controller optimization, the implementation uses several auxiliary regularizers and stabilizers that are not part of the main algorithmic description. An edge-activation cost
  penalizes every edge that is actually executed, encouraging sparse graphs. A blocked-edge penalty penalizes an edge that was proposed by the controller but later rejected by legality
  checks or post-processing before execution. An effect-size threshold suppresses any proposed edge whose learned causal-force vector has norm below a fixed threshold, treating it as
  too weak to contribute to state propagation; an effect-size penalty then penalizes selecting such weak edges in the first place. The controller objective also includes an entropy
  bonus, which encourages exploration by discouraging premature collapse to deterministic policies. During training, both controllers share a temperature schedule that anneals linearly
  from 2.0 to 1.0, whereas during validation and evaluation the temperature is fixed to 1.0 and controller decisions are deterministic by default. Temporal-difference learning further
  uses slowly updated target controllers via Polyak averaging \citep{doi:10.1137/0330046}, and stored Q-values are clipped to the range [-10, 10] for stability. In the default configuration, the edge-
  activation cost is 0.1, the blocked-edge penalty is 0.1, the effect-size threshold is 0.1, the effect-size penalty is 0.15, and the entropy-bonus coefficient is 0.1. Importantly, an
  edge that is proposed but later blocked is still kept in training as a negative example rather than being discarded.

  A further implementation detail concerns how sparse acyclic interaction graphs are enforced during recurrent rollout. The modeled scene entities and the pairwise forces between them
  are referred to as endogenous objects and endogenous forces. Separate exogenous objects and exogenous forces represent influences originating outside that modeled object set. In the
  default configuration there is a single exogenous object, intended to represent the surrounding environment or boundary effects, such as walls. Graph construction proceeds over
  recurrent steps using a two-tier frontier, where a frontier is the set of object nodes currently eligible to participate in the next wave of interaction propagation. The endogenous
  frontier is defined by sink nodes of currently executed endogenous edges, where a sink is the object receiving the effect of a directed interaction. If no such sink exists at the
  beginning of rollout, the object directly targeted by the action is used as the initial seed frontier. Endogenous edge proposals are restricted to object pairs touching the current
  endogenous frontier, and their directions are further constrained so that frontier nodes act as sources for subsequent diffusion rather than immediate sinks. After an object is
  directly intervened on by the action, incoming forces to that object are suppressed for the next recurrent step, so the intervened object temporarily seeds propagation without itself
  being immediately re-entered as a sink. The exogenous frontier is cumulative: it contains all endogenous objects that have ever been intervention targets or sinks in earlier
  recurrent steps. Exogenous edges may target only this cumulative set, and only when the endogenous frontier is empty or no endogenous edge is active. Directed-acyclic-graph (DAG)
  structure is enforced at two levels: first, candidate endogenous edges whose admissible directions would introduce a cycle are masked before activation; second, after direction
  selection, any direction that would violate acyclicity is flipped to the valid alternative when one exists, and otherwise the edge is deactivated. A final sequential reachability
  update ensures that the executed directed graph remains acyclic throughout the rollout.

\section{Plots}
\label{sec:plots}

\begin{figure}[H]
    \begin{center}
        \centerline{\includegraphics[width=\textwidth]{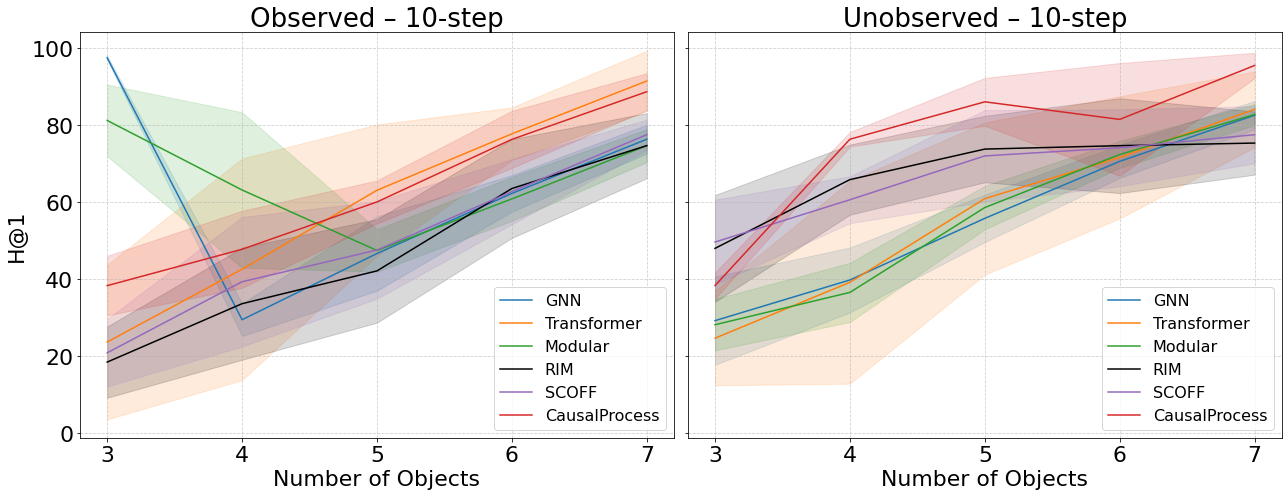}}
    \end{center}
    \caption{Prediction metric vs number of objects for 10-steps.}
    \label{fig:h@1-10-step-other-baselines}
\end{figure}

\begin{figure}[H]
    \begin{center}
        \centerline{\includegraphics[width=\textwidth]{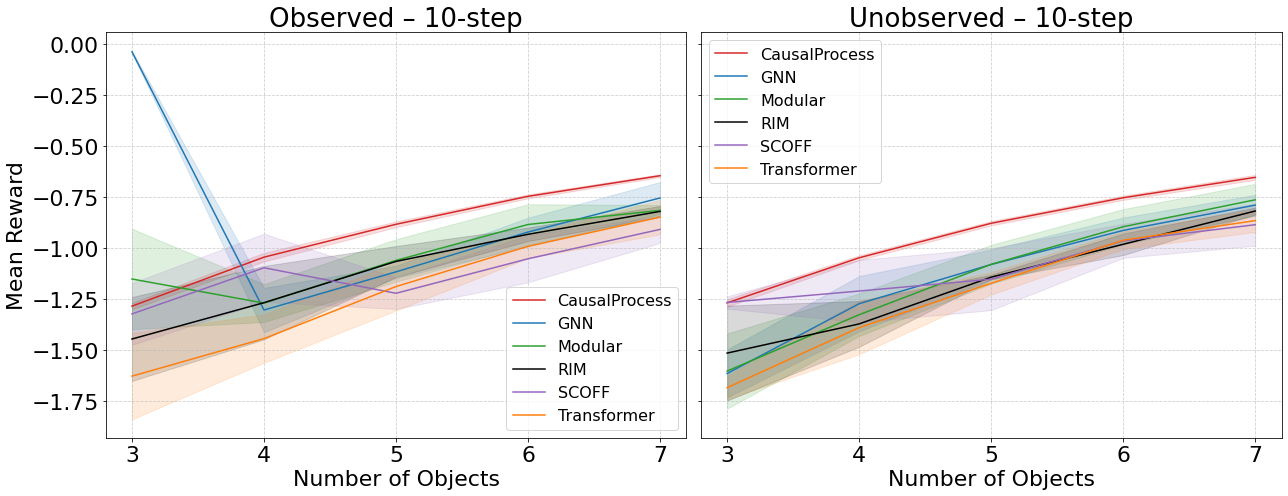}}
    \end{center}
    \caption{\emph{Downstream RL results over number of objec}. Mean reward vs number of objects. All results are the average of 10 seeds}
    \label{fig:mean-reward-other-basleines}
    \vspace{-1.5em}
\end{figure}

\begin{figure}[H]
    \begin{center}
        \centerline{\includegraphics[width=\textwidth]{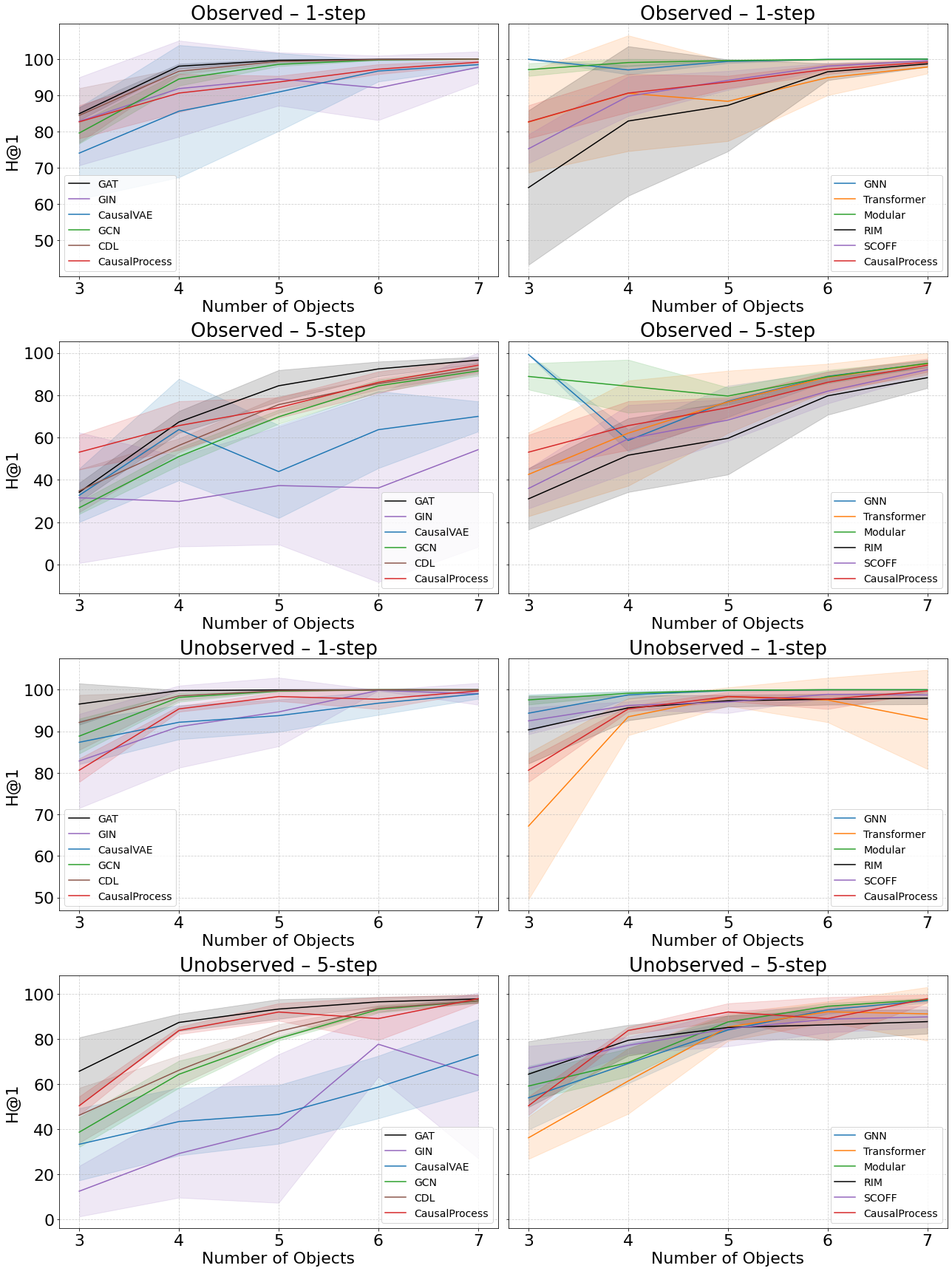}}
    \end{center}
    \caption{Prediction metric vs number of objects after 1 and 5 steps (average of 10 seeds).}
    \label{fig:h@1-1-5-steps}
\end{figure}

\begin{figure}[H]
    \begin{center}
        \centerline{\includegraphics[width=\textwidth]{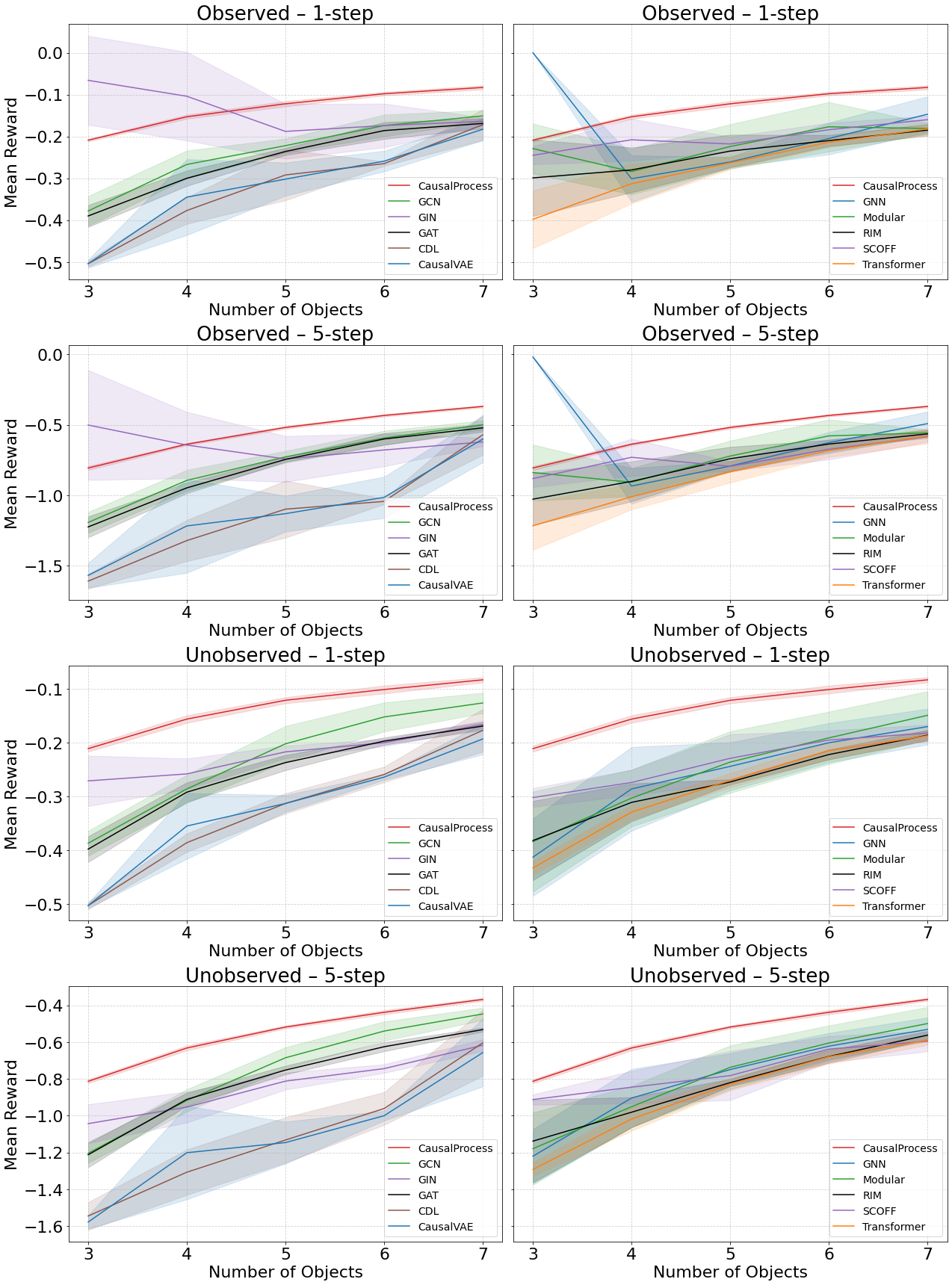}}
    \end{center}
    \caption{Mean reward vs number of objects after 1 and 5 steps (average of 10 seeds).}
    \label{fig:mean-reward-1-5-steps}
\end{figure}

\newpage

\section{Ablation Study}
\label{app:ablation}

The CPM combines (i) a structured latent factorization, and (ii) an RL-based procedure for selecting
sparse, time-varying causal edges. To isolate which components are essential for performance, we
conduct targeted ablations of the three most central design choices: learned reward shaping, learned
causal controllers, and the structured C/P/M representation.

\paragraph{Setup.}
All ablations are evaluated in the \textbf{Observed} setting with \textbf{3 objects}, averaged over
\textbf{2 random seeds}. We report both (i) prediction accuracy via Hits@1 (H@1; Sec.~\ref{subsec:prediction}) and
(ii) downstream model-based RL performance via mean environment reward (Sec.~\ref{subsec:downstream_rl}), across
rollout horizons $K\in\{1,5,10\}$.
We report \emph{deltas} computed as
$\Delta = \text{(ablated)} - \text{(full CPM)}$,
so negative values indicate degradation relative to the full model.
For each ablation, we keep the training pipeline and hyperparameters fixed and disable exactly one
component, leaving all others intact.

\paragraph{Reward ablated (zero reward shaping).}
We replace the learned reward models $R_O$ and $R_{O\leftrightarrow F}$ (Eq.~\ref{eq:rewards}) with constant
zero outputs. This removes the inverse-RL reward shaping signal that otherwise trains the
controllers to prefer graph decisions that reduce prediction error.

\paragraph{Controller ablated (random graph decisions).}
We replace the learned controller policies $\pi_O$ and $\pi_{O\leftrightarrow F}$ (Sec.~\ref{model:controller}) with
uniform random edge selection (while keeping the same structural constraints, e.g., pairwise
interactions / mirroring when applicable). This removes the model’s ability to adaptively select
sparse causal edges based on the current state.

\paragraph{C/P/M regions ablated (fused latent).}
We remove the structured factorization into Causal relevance (C), Control relevance (P), and
Mutability (M) (Sec.~\ref{model:struc_rep}) and instead use a single undifferentiated latent vector with the same
total dimensionality. This ablates the hard architectural routing bias that forces semantically
distinct subspaces.

\paragraph{Results and discussion.}
Table~\ref{tab:ablation} summarizes the
results. All ablations consistently degrade both prediction and downstream planning, and the effect
\emph{amplifies strongly with rollout horizon}. At $K=1$, the H@1 drops are small ($\leq 0.27$
percentage points), while at $K=10$ they become substantial (up to $-13.01$ H@1 points and
$-1.08$ reward). This pattern suggests that the components are particularly important in the
long-horizon regime where compounding error is the dominant failure mode.

Among the components, the structured \textbf{C/P/M factorization} is most critical at long horizons
($\Delta$H@1 $=-13.01$ at $K=10$; $\Delta$Reward $=-1.08$), consistent with its intended role of
separating immutable vs.\ mutable and causally relevant vs.\ irrelevant factors during multi-step
rollouts. The \textbf{reward shaping} ablation yields the second-largest long-horizon degradation in
H@1 ($-10.63$ at $K=10$), supporting the claim that learned rewards are important for training the
controllers toward causally useful graphs. The \textbf{random-controller} ablation shows smaller but
still meaningful degradation ($-8.45$ H@1 at $K=10$), which we note may be partially understated
in the 3-object setting because the space of possible object pairs is small; with more objects, random
edge selection becomes less likely to pick the correct interacting pair.

\begin{table}[t]
\centering
\caption{\textbf{Ablation results (Observed, 3 objects, mean over 2 seeds).}
$\Delta$ values are computed as \emph{ablated $-$ full CPM}. Negative indicates worse.}
\label{tab:ablation}
\begin{tabular}{l c c c c c c}
\toprule
& \multicolumn{3}{c}{$\Delta$ H@1 (\% points)} & \multicolumn{3}{c}{$\Delta$ Reward} \\
\cmidrule(lr){2-4} \cmidrule(lr){5-7}
Ablation & 1-step & 5-step & 10-step & 1-step & 5-step & 10-step \\
\midrule
Reward ablated & $-0.27$ & $-3.97$ & $-10.63$ & $-0.20$ & $-0.61$ & $-0.74$ \\
Controller ablated & $-0.17$ & $-3.52$ & $-8.45$ & $-0.09$ & $-0.21$ & $-0.24$ \\
C/P/M ablated & $-0.19$ & $-4.12$ & $-13.01$ & $-0.24$ & $-0.85$ & $-1.08$ \\
\bottomrule
\end{tabular}
\end{table}



\section{Graph Recovery and Structured Representation Analysis (Observed 3-object setting)}
\label{app:graph_and_repr}

\subsection{Linear probing of the structured $(C,P,M)$ subspaces}
\label{app:graph_and_repr:probing}

We test whether \emph{object location} is preferentially encoded in
the \emph{mutable} ($M{=}1$) subspaces, consistent with the intended semantics of $M$ (Sec.~4.1.1):
positions change over time, whereas immutable properties should not.

\paragraph{Protocol.}
For each trained CPM (two seeds), we freeze all model parameters and extract the
object latents $O_i^t$, which are partitioned into the eight sub-vectors corresponding to all
$(C,P,M)\in\{0,1\}^3$:
\[
O_i^t = \bigoplus_{C,P,M\in\{0,1\}} O_i^{t,CPM}.
\]
For each sub-vector (e.g., $O_i^{t,CPM}$, $O_i^{t,CM}$, \dots, $O_i^{t,\varnothing}$), we train a
linear regressor to predict object location and report held-out test $R^2$. We use the same
train/test split across seeds and report mean $\pm$ std over seeds.

\paragraph{Result.}
Location is almost perfectly decodable from all subspaces with $M{=}1$ (CPM/CM/PM/M),
with $R^2 \approx 0.99$, while decodability collapses in $M{=}0$ subspaces (CP/C/P/$\varnothing$),
with $R^2 \approx 0.08{-}0.14$ on average (
Table~\ref{tab:appendix_location_probe}). This supports that the model routes positional information
into the \textbf{mutable} channel, aligning with the inductive bias described in Sec.~4.1.1.


\begin{table}[t]
\centering
\begin{tabular}{l c}
\toprule
Region & Location $R^2$ (mean $\pm$ std) \\
\midrule
$M$   & 0.991 $\pm$ 0.003 \\
$PM$  & 0.991 $\pm$ 0.003 \\
$CM$  & 0.991 $\pm$ 0.003 \\
$CPM$ & 0.990 $\pm$ 0.003 \\
\midrule
$\varnothing$ & 0.141 $\pm$ 0.137 \\
$C$           & 0.115 $\pm$ 0.087 \\
$CP$          & 0.091 $\pm$ 0.088 \\
$P$           & 0.078 $\pm$ 0.084 \\
\bottomrule
\end{tabular}
\caption{
Linear probing of CPM object subspaces for \textbf{location}.
  We train a linear regressor to predict object position from each $(C,P,M)$ sub-vector.
  The combined panel reports mean $\pm$ std $R^2$ over two seeds.
  Location is almost perfectly decodable from all subspaces with $M{=}1$ (CPM/CM/PM/M),
  but not from $M{=}0$ subspaces (CP/C/P/$\varnothing$), indicating that positional information
  is routed into the \textbf{mutable} channel. Values are test $R^2$
(mean $\pm$ std over two seeds).
}
\label{tab:appendix_location_probe}
\end{table}

\subsection{Graph recovery against the ground-truth interaction graph}
\label{app:graph_and_repr:graph_recovery}

To provide a quantitative measure of causal graph discovery, we compare the learned directed
interaction graph to the simulator's ground-truth interaction graph in the observed 3-object setting.

\paragraph{Ground truth.}
In the \citet{DBLP:conf/nips/KeDMGLBRMBP21} environment used in our experiments, interactions are sparse and directed:
the only force is pushing (double-pushes are disallowed), and only heavier objects can push lighter
ones. Thus, when an interaction occurs, the ground-truth directed edge is always
\emph{heavy$\rightarrow$light}. (In the observed setting, weight corresponds to color intensity.)

\paragraph{Predicted graph and metrics.}
At each timestep, CPM predicts a sparse directed interaction graph. We threshold edge activations
at $0.5$ to obtain a discrete predicted graph and compute:
(i) Precision/Recall/F1 for edge presence vs.\ ground truth, and
(ii) \textbf{Direction validity} = fraction of predicted edges that follow the correct
heavy$\rightarrow$light direction.
We report metrics both over \emph{all steps} (including non-interaction timesteps) and over
\emph{interaction steps only} (timesteps where the ground-truth graph is non-empty).

\paragraph{Result.}
On interaction timesteps, CPM recovers directed interaction edges with substantially higher precision
and predicts the correct heavy$\rightarrow$light direction in the large majority of predicted edges
(Table~\ref{tab:appendix_graph_recovery}). Over all timesteps, precision is lower due to false
positives on non-interaction steps, suggesting that sparsity calibration is an important direction for
future improvement.

\begin{table}[t]
\centering
\begin{tabular}{l c c c c}
\toprule
Evaluation set & Precision & Recall & F1 & Direction validity \\
\midrule
All steps
& 0.084 $\pm$ 0.010
& 0.403 $\pm$ 0.059
& 0.139 $\pm$ 0.018
& 0.738 $\pm$ 0.099 \\
Interaction steps only
& 0.544 $\pm$ 0.050
& 0.403 $\pm$ 0.059
& 0.463 $\pm$ 0.057
& 0.853 $\pm$ 0.088 \\
\bottomrule
\end{tabular}
\caption{
Graph recovery in the observed 3-object setting (mean $\pm$ std over two seeds).
Direction validity measures the fraction of predicted edges that follow the correct
heavy$\rightarrow$light direction.
}
\label{tab:appendix_graph_recovery}
\end{table}

\subsection{Visualization of the learned directed graph}
\label{app:graph_and_repr:viz}

\begin{figure}[t]
  \centering
  \begin{minipage}{0.32\linewidth}
    \centering
    \includegraphics[width=\linewidth]{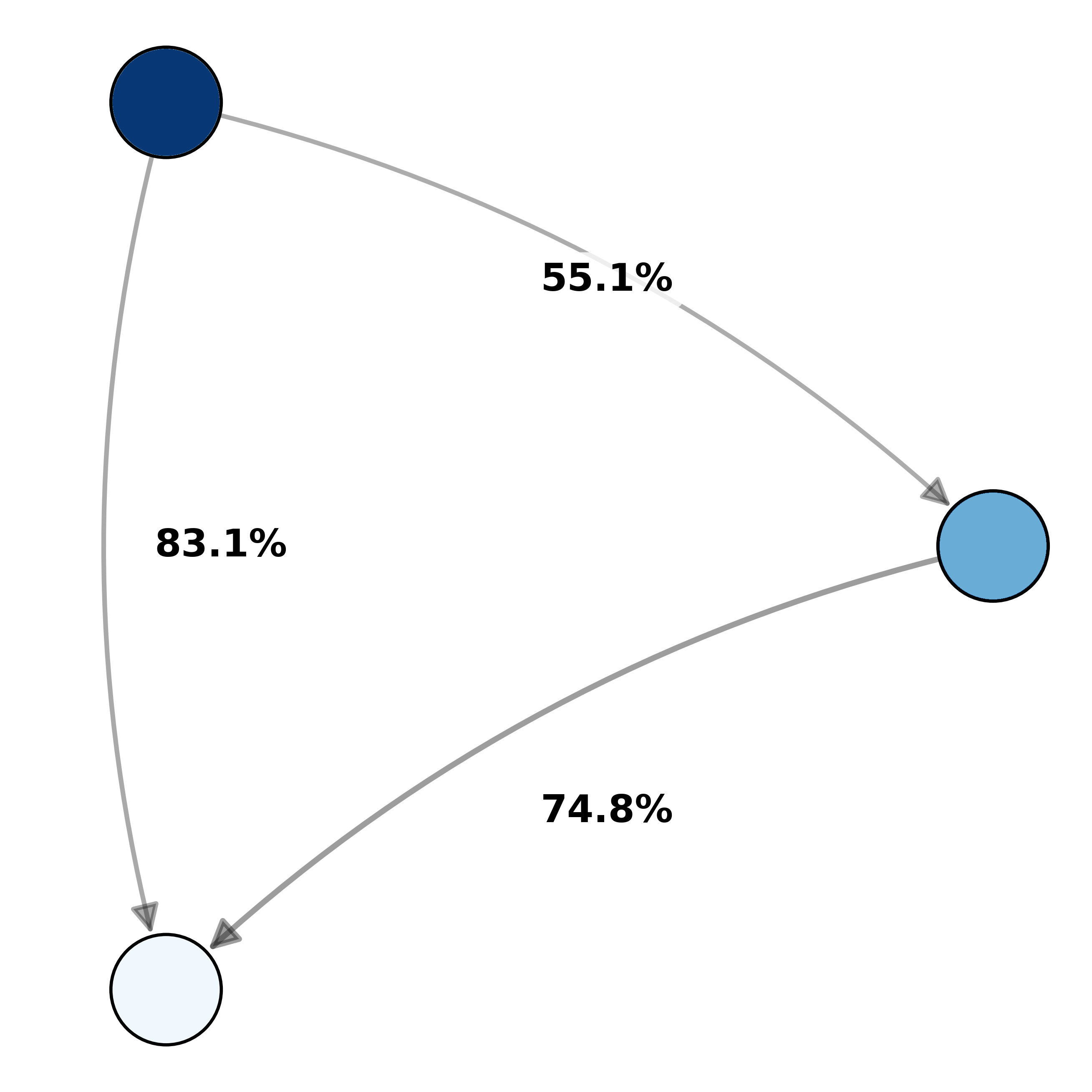}
    \caption{1\textsuperscript{st} Learned DAG.}
  \end{minipage}
  \begin{minipage}{0.32\linewidth}
    \centering
    \includegraphics[width=\linewidth]{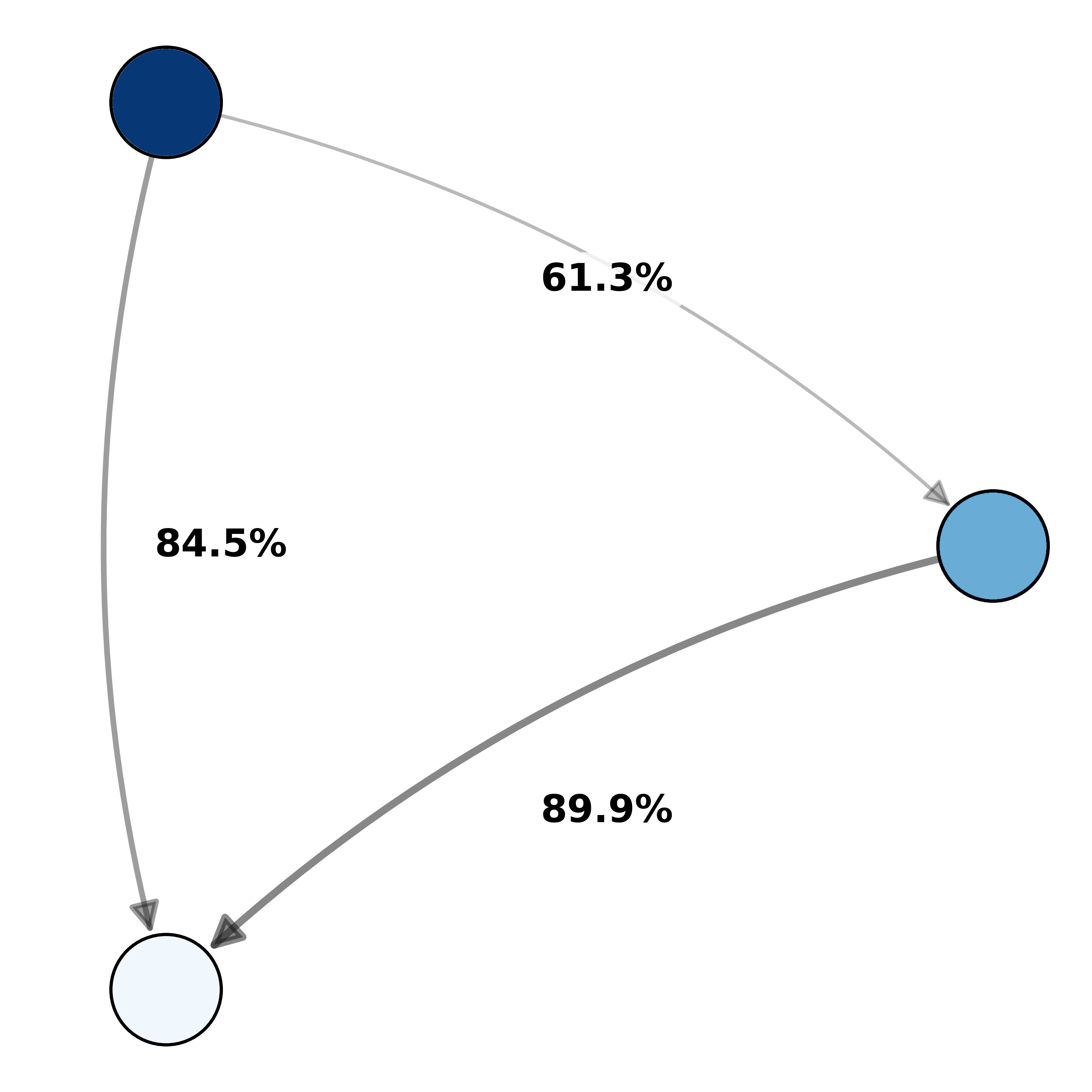}
    \caption{2\textsuperscript{nd} Learned DAG.}
  \end{minipage}
  \begin{minipage}{0.32\linewidth}
    \centering
    \includegraphics[width=\linewidth]{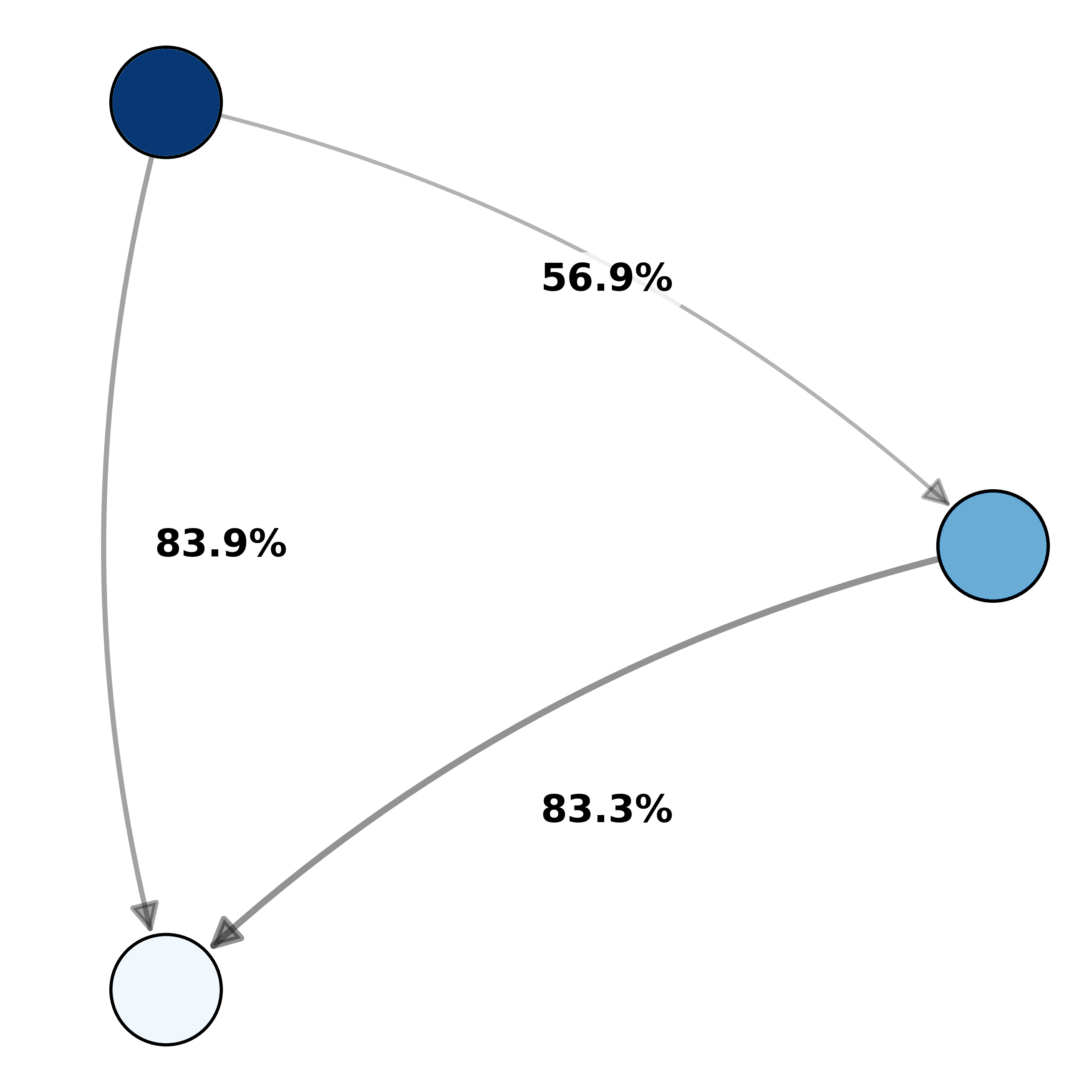}
    \caption{Average}
  \end{minipage}
  \caption{
  Learned directed interaction graph in the observed 3-object setting.
  For each pair of objects $\{i,j\}$ we average predicted edge probabilities $p(i\rightarrow j)$ over the
  dataset, draw the arrow in the more probable direction, and label it with the directional share
  $\frac{p(i\rightarrow j)}{p(i\rightarrow j)+p(j\rightarrow i)}$.
  This visualization qualitatively matches the quantitative directionality reported in
  Table~\ref{tab:appendix_graph_recovery}.
  }
  \label{fig:appendix_learned_graph}
\end{figure}

To provide an interpretable summary of the learned graph structure, we average the model's
predicted edge probabilities $p(i\rightarrow j)$ over the evaluation dataset.
For each unordered pair $\{i,j\}$, we draw the arrow in the more probable direction and label it with
the \emph{directional share}:
\[
\frac{p(i\rightarrow j)}{p(i\rightarrow j)+p(j\rightarrow i)} \times 100\%.
\]
Fig.~\ref{fig:appendix_learned_graph} shows per-seed graphs and their average. Nodes are colored
by object weight (darker indicates heavier), so heavy$\rightarrow$light corresponds to dark$\rightarrow$light.

\end{document}